\documentclass[preprint,12pt]{elsarticle}

\usepackage{xcolor}
\usepackage{algorithm}
\usepackage{algpseudocode}
\usepackage{multirow}

\usepackage{booktabs}
\usepackage{amsmath,amssymb,amsfonts}
\usepackage{graphicx}
\usepackage{bm}

\newcommand{\bb}{\boldsymbol{b}}

\newcommand{\bu}{\boldsymbol{u}}

\newcommand{\bz}{\boldsymbol{z}}

\newtheorem{remark}{Remark}
\journal{Journal of Computational Physics}
\usepackage{comment}
\begin{document}
	
	\begin{frontmatter}
		\title{A Hybrid Physics-informed holomorphic neural
			networks Method}
		
		\author[1]{Zhijie Su}
		\ead{zhijiesu164@163.com}
		
		\author[1]{Qiaolin He\corref{cor1}}
		\ead{qlhejenny@scu.edu.cn}
		
		\author[2]{Zhiping Mao\corref{cor1}}
		\ead{zmao@eitech.edu.cn}
		
		\author[1]{Shiquan Zhang}
		\ead{shiquanzhang@scu.edu.cn}
		
		\cortext[cor1]{Corresponding authors.}
		
		\affiliation[1]{organization={Department of Mathematics, Sichuan University},
			addressline={No. 24, Section 1, South Yihuan Road},
			city={Chengdu},
			postcode={610065},
			country={China}}
		
		\affiliation[2]{organization={Eastern Institute of Technology, Ningbo},
			addressline={No. 2911, Haijiang Avenue, Zhenhai District},
			city={Ningbo},
			postcode={315200},
			country={China}}
		
		%% Abstract
		\begin{abstract}
			Physics-informed holomorphic neural networks (PIHNNs) exploit complex-analytic representations to satisfy homogeneous PDEs by construction, but their extension to nonhomogeneous problems is nontrivial. We propose a hybrid physics-informed holomorphic neural network (HPIHNN) framework based on a decomposition of the solution into particular and homogeneous components. The particular component is computed by a spectral method on a simple extended domain, while the homogeneous component is approximated by a PIHNN trained only on the boundary with modified boundary data. The resulting formulation avoids meshing complex geometries and retains the dimensionality-reduction advantage of holomorphic representations. The method is applied to nonhomogeneous Poisson, Stokes, biharmonic, linear elasticity, and Helmholtz equations. Numerical experiments covering complex geometries, corner singularities, and nearly incompressible regimes demonstrate that the proposed method accurately resolves the corresponding solutions and, in the considered tests, generally achieves lower errors and shorter training times than standard PINNs. The results indicate that coupling spectral particular solutions with holomorphic neural approximations provides an efficient approach to nonhomogeneous PDEs on complex domains.
		\end{abstract}
		
		
		
		%% Keywords
		\begin{keyword}
			Physics-informed holomorphic neural networks\sep Nonhomogeneous partial differential equations \sep Spectral methods \sep Linear superposition\sep  Complex geometries
		\end{keyword}
		
	\end{frontmatter}

	\section{Introduction} \label{sec1}
	
	Partial differential equations (PDEs), including the Poisson, Stokes, linear elasticity, and biharmonic equations, arise in the mathematical modeling of potential fields, viscous flows, elastic deformation, and thin-plate bending \cite{Stokes,Possion,Elasticity,Biharmonic}. Their accurate numerical solution in geometrically complex domains remains a fundamental problem in classical computational mathematic discretization techniques, such as finite difference, finite element, and finite volume methods, provide reliable and well-established frameworks, but their implementation in complex geometries may involve nontrivial mesh generation, boundary treatment, and high-order discretization \cite{FDM,FEM,FVM}. For fourth-order problems such as the biharmonic equation, conforming finite element formulations additionally require $C^1$-continuous approximation spaces, while mixed or nonconforming formulations typically introduce auxiliary variables or additional degrees of freedom \cite{FEM2}. Spectral methods can achieve very high accuracy for sufficiently smooth solutions, but their extension to irregular domains generally requires specialized geometric treatments \cite{Spectral}. These considerations motivate the development of flexible meshless solvers for PDEs posed on complex domains.
	
	In recent years, neural-network-based PDE solvers have emerged as an alternative meshless paradigm. Among them, physics-informed neural networks (PINNs) \cite{PINN} incorporate the governing equations and associated initial or boundary conditions into the training objective, with the required derivatives evaluated by automatic differentiation (AD). This formulation avoids an explicit spatial mesh and has consequently been applied to a broad range of PDEs, including Poisson problems \cite{PINN-Possion}, incompressible Stokes flow \cite{PINN-Stokes1}, and linear elasticity \cite{PINN-Elasticity1,PINN-Elasticity2}. Despite this flexibility, standard PINNs may still suffer from high training cost, optimization difficulties, and reduced accuracy for high-order operators, complex geometries, or solutions with localized features, motivating the development of more structured neural formulations.

	Considerable effort has been devoted to improving the trainability and
	scalability of PINNs. Adaptive activation functions have been introduced to accelerate convergence \cite{Adp-PINN}, while adaptive loss-weighting strategies have been developed to alleviate gradient imbalance during training \cite{Wang1,Wang2}. Domain decomposition approaches, such as eXtended Physics-Informed Neural Networks (XPINNs) \cite{XPINN}, and discrete PINN formulations \cite{fPINN} have further extended the applicability of PINNs to large-scale and complex problems. Nevertheless, several challenges remain. PINN training involves a highly nonconvex optimization problem and may exhibit slow convergence and
	sensitivity to optimizer and loss-weight selection \cite{PINN-fail,PINN-compare}. For high-order PDEs, repeated automatic differentiation substantially increases computational and memory costs \cite{DeepXDE,PINN-highorder}. Moreover, the boundary conditions in standard PINNs are typically imposed through penalty terms and therefore not exactly satisfied unless specially constructed hard-constraint
	representations are employed \cite{PINN-boundary,All-PINN}. These
	limitations motivate neural formulations that incorporate more analytical
	structure of the underlying PDEs into the approximation space.
	
	An alternative direction is to exploit complex-valued and holomorphic
	representations. Complex-valued neural networks (CVNNs) have been studied
	since the 1990s \cite{CVNN1,CVNN2} and continue to attract interest
	\cite{CVNN-NEW}, although most developments have remained primarily
	data-driven. For PDE approximation, Ghosh et al. introduced a holomorphic
	neural network (HNN) \cite{HNN} for the two-dimensional Laplace equation.
	This idea was subsequently extended to physics-informed holomorphic neural
	networks (PIHNNs) \cite{PIHNN}, where the Kolosov--Muskhelishvili
	representation \cite{KM} is incorporated to solve two-dimensional
	homogeneous linear elasticity problems on complex geometries. By embedding
	the analytical structure of the governing equations into the network
	representation, these methods satisfy the homogeneous PDEs by construction
	and reduce training essentially to the enforcement of boundary conditions.
	This dimensionality reduction avoids interior residual evaluation and is
	particularly attractive for problems posed on geometrically complex
	domains.

	Existing holomorphic neural formulations are primarily restricted to
	homogeneous governing equations. To extend this approach to nonhomogeneous
	problems, we propose a hybrid physics-informed holomorphic neural network
	(HPIHNN) framework based on linear superposition. The solution is decomposed
	into a particular component, computed by a spectral method on a simple
	extended domain, and a homogeneous component represented by a PIHNN with
	modified boundary data. This construction preserves the boundary-only
	training structure of PIHNNs while avoiding interior residual evaluation
	and the associated AD cost. 
	
	The proposed method is evaluated on nonhomogeneous Poisson, Stokes, biharmonic, linear elasticity, and Helmholtz problems involving complex geometries, corner singularities, and nearly incompressible regimes.
	Numerical comparisons with standard PINNs are used to assess both accuracy
	and computational efficiency. The remainder of the paper is organized as follows.
	Section~\ref{sec2} reviews HNNs and PIHNNs \cite{HNN,PIHNN}.
	Section~\ref{sec3} introduces the HPIHNN framework.
	Section~\ref{sec4} presents the numerical experiments, and
	Section~\ref{sec5} concludes the paper.
	
	\section{Physics-informed holomorphic neural networks}
	\label{sec2}

	We briefly review the PIHNN framework \cite{HNN,PIHNN}. Consider
	\begin{equation}\label{eq:PDE1}
		\mathcal A u=0 \quad \text{in } \Omega,
		\qquad
		\mathcal B u=g \quad \text{on } \partial\Omega.
	\end{equation}
	For $\Omega\subset\mathbb R^2$, we identify $(x,y)$ with
	$z=x+\mathrm{i}y$ and define
	$\Omega_{\mathbb C}=\{x+\mathrm{i}y:(x,y)\in\Omega\}$.
	A complex-valued neural network is represented by
	\begin{equation}\label{eq:NN}
		\mathcal N_\theta
		=
		L_k\circ\sigma\circ L_{k-1}\circ\cdots\circ\sigma\circ L_1,
		\qquad
		L_j(\bz)=W_j\bz+\bb_j.
	\end{equation}
	If $\sigma$ is holomorphic, then $\mathcal N_\theta$ is holomorphic as a
	composition of holomorphic mappings. PIHNN combines this representation
	with problem-specific complex-variable formulations so that the homogeneous
	governing equation is satisfied by construction. The neural-network
	optimization can therefore be reduced to the enforcement of boundary
	conditions.

	Existing holomorphic neural formulations \cite{HNN,PIHNN} exploit
	problem-specific complex-variable representations to satisfy homogeneous
	governing equations by construction. The neural-network parameters are
	therefore determined primarily by minimizing boundary residuals.
	
	For the homogeneous Laplace equation, the solution in \cite{HNN} is represented as
	\begin{equation}\label{eq:Lap}
		u(x,y)=\operatorname{Re}\bigl(\mathcal N_\theta(z)\bigr),
		\qquad z=x+\mathrm{i}y,
	\end{equation}
	where the holomorphicity of $\mathcal N_\theta$ implies
	$\Delta u=0$. 
	The loss function in \cite{HNN} is defined according to the Dirichlet boundary condition imposing on the boundary $\partial \Omega$. %adopted from Eq.~(7) of \cite{HNN}, is defined as
	%	\begin{equation}\label{eq:Loss_lapulace}
		%		\mathcal L=\left\|g-\operatorname{Re}\bigl(\mathcal N_\theta(z)\bigr)\right\|_2.
		%	\end{equation}
	
	For two-dimensional homogeneous linear elasticity, PIHNN
	\cite{PIHNN} employs the Kolosov--Muskhelishvili representation
	\cite{KM}. In terms of two holomorphic potentials $\varphi,\psi\in\mathcal H(\Omega_{\mathbb C})$, the stress and displacement fields are given by
	\begin{equation}\label{eq:KM}
		\begin{aligned}
			\sigma_{xx}
			&=\operatorname{Re}\!\left(
			2\varphi'(z)-\overline z\,\varphi''(z)-\psi'(z)
			\right),\\
			\sigma_{yy}
			&=\operatorname{Re}\!\left(
			2\varphi'(z)+\overline z\,\varphi''(z)+\psi'(z)
			\right),\\
			\sigma_{xy}
			&=\operatorname{Im}\!\left(
			\overline z\,\varphi''(z)+\psi'(z)
			\right),\\
			u_x
			&=\frac{1}{2\mu}
			\operatorname{Re}\!\left(
			\gamma\varphi(z)
			-z\,\overline{\varphi'(z)}
			-\overline{\psi(z)}
			\right),\\
			u_y
			&=\frac{1}{2\mu}
			\operatorname{Im}\!\left(
			\gamma\varphi(z)
			-z\,\overline{\varphi'(z)}
			-\overline{\psi(z)}
			\right),
		\end{aligned}
	\end{equation}
	where $\mu$ denotes the shear modulus,
	respectively, and $\gamma$ is the corresponding Kolosov constant, the loss function for the linear elasticity problem is defined by Eq.~(6) in \cite{PIHNN}. %is the shear modulus and
	% \begin{equation}\label{eq:kolosov_constant}
		% \gamma=
		% \begin{cases}
			% \dfrac{3-\nu}{1+\nu}, & \text{plane stress},\\[2mm]
			% 3-4\nu, & \text{plane strain},
			% \end{cases}
		% \end{equation}
	%with $\nu$ denoting Poisson's ratio.
	
	Because these representations satisfy the corresponding homogeneous
	governing equations analytically, PIHNN training can be restricted to
	boundary collocation points. Compared with standard PINNs, this eliminates
	interior PDE-residual evaluation and reduces the dimension of the
	collocation problem, which can be particularly advantageous for complex
	geometries. However, the formulations in \cite{HNN,PIHNN} are constructed
	for homogeneous governing equations and do not directly accommodate
	nonhomogeneous source terms. This restriction motivates the hybrid
	formulation developed in the next section.

	\section{A Hybrid Physics-informed holomorphic neural networks Method
	}
	\label{sec3}
	% In this section, to address more general non-homogeneous cases, we innovate upon the PIHNN framework by leveraging the principle of superposition and propose a novel method termed : A Hybrid Physics-Informed Holomorphic Neural Networks Method (HPIHNN). In Section \ref{sec3.1}, we introduce the general framework of HPIHNN for solving non-homogeneous problems. In Section \ref{sec3.2}, we discuss several classes of problems whose homogeneous fundamental solutions can be represented by holomorphic functions, along with their corresponding analytical expressions.
	In this section, we extend the PIHNN framework to nonhomogeneous linear PDEs through a superposition-based decomposition and introduce the hybrid physics-informed holomorphic neural network (HPIHNN) method. Section~\ref{sec3.1} presents the general formulation, and
	Section~\ref{sec3.2} discusses several representative PDEs whose homogeneous solutions admit holomorphic representations.
	\subsection{HPIHNN formulation}
	\label{sec3.1}
	% Consider the general representation of a linear partial differential equation (PDE) defined on a computational domain $\Omega$:
	% \begin{equation}
		% 	\begin{cases} 
			% 		\mathcal{A}u=f, & \text{in } \Omega, \\ 
			% 		\mathcal{B}u=g, & \text{on } \partial\Omega, 
			% 	\end{cases}
		% 	\label{eq:PDE}
		% \end{equation}
	% where $\mathcal{A}$ and $\mathcal{B}$ represent the differential operator and the boundary operator, respectively, both of which are linear. Utilizing the principle of superposition, we decompose the solution $u$ of the original problem \eqref{eq:PDE} as:
	% \begin{equation}
		% 	u = u_1 + u_2,
		% 	\label{eq:u=u1+u2}
		% \end{equation}
	% such that the following conditions are satisfied:
	% \begin{equation}
		% 	\begin{cases}
			% 		\mathcal{A}u_1=f, & \text{in } \Omega, \\ 
			% 		\mathcal{B}u_1=g_{u_1}, & \text{on } \partial\Omega, 
			% 	\end{cases}
		% 	\label{eq:Superposition principle1}
		% \end{equation}
	% and
	% \begin{equation}
		% 	\begin{cases}
			% 		\mathcal{A}u_2=0, & \text{in } \Omega, \\ 
			% 		\mathcal{B}u_2=g-g_{u_1}, & \text{on } \partial\Omega.
			% 	\end{cases}
		% 	\label{eq:Superposition principle2}
		% \end{equation}
	Consider the nonhomogeneous linear boundary-value problem \eqref{eq:PDE},
	\begin{equation}\label{eq:PDE}
		\begin{cases}
			\mathcal A u=f, & \text{in } \Omega,\\
			\mathcal B u=g, & \text{on } \partial\Omega,
		\end{cases}
	\end{equation}
	where $\mathcal A$ and $\mathcal B$ are linear differential and boundary
	operators, respectively. By linear superposition, we decompose
	\begin{equation}\label{eq:u=u1+u2}
		u=u_p+u_h,
	\end{equation}
	where the particular component $u_p$ satisfies
	\begin{equation}\label{eq:Superposition principle1}
		\begin{cases}
			\mathcal A u_p=f, & \text{in } \Omega,\\
			\mathcal B u_p=g_p, & \text{on } \partial\Omega,
		\end{cases}
	\end{equation}
	and the homogeneous component $u_h$ satisfies
	\begin{equation}\label{eq:Superposition principle2}
		\begin{cases}
			\mathcal A u_h=0, & \text{in } \Omega,\\
			\mathcal B u_h=g-g_p, & \text{on } \partial\Omega.
		\end{cases}
	\end{equation}
	Here,
	\[
	g_p:=%\left.\mathcal B \widetilde{u}_p\right|_{\partial\Omega}
	\left.\mathcal B u_p\right|_{\partial\Omega}
	\]
	denotes the boundary data induced by the particular solution.

	The homogeneous correction problem \eqref{eq:Superposition principle2}
	is solved by the PIHNN method reviewed in Section~\ref{sec2}. Its
	boundary data depend on the trace of the particular solution on
	$\partial\Omega$. To construct the particular component $u_p$ in \eqref{eq:Superposition principle1} efficiently, we
	embed the original domain $\Omega$ into a larger regular domain
	$\widetilde{\Omega}\supset\Omega$ and solve the following extending problem
	\begin{equation}\label{eq:Superposition principle3}
		\begin{cases}
			\mathcal A\widetilde{u}_p=\widetilde f,
			& \text{in } \widetilde{\Omega},\\
			\widetilde{\mathcal B}\widetilde{u}_p=\widetilde g_p,
			& \text{on } \partial\widetilde{\Omega},
		\end{cases}
	\end{equation}
	where $\widetilde f|_{\Omega}=f$, and
	$\widetilde{\mathcal B}$ and $\widetilde g_p$ specify the boundary conditions imposed on the extended, geometrically simple domain $\partial\widetilde{\Omega}$. For example,
	periodic conditions on opposing faces are natural for Fourier spectral
	methods, whereas homogeneous Dirichlet conditions are one common choice
	for Chebyshev spectral methods. The particular solution on the original domain $\Omega$ is then defined by
	\[
	u_p:=\widetilde u_p|_{\Omega},
	\qquad
	g_p:= \left.\mathcal B \widetilde u_p\right|_{\partial\Omega} = \left.\mathcal B u_p\right|_{\partial\Omega}.
	\]
	Since $\widetilde{\Omega}$ is chosen to have a simple geometry,
	\eqref{eq:Superposition principle3} can be solved using standard numerical methods. In this work, we employ a spectral method.
	%The homogeneous component $u_h$ satisfies
	%\[
	%\mathcal A u_h=0 \quad \text{in } \Omega,
	%\qquad
	%\mathcal B u_h=g-g_p \quad \text{on } \partial\Omega,
	%\]
	%and is approximated by a PIHNN. 
	Let $\{(x_i,y_i)\}_{i=1}^N\subset\partial\Omega$ be boundary collocation points,
	with $z_i=x_i+\mathrm{i}y_i$. Using holomorphic neural potentials
	$\Psi$ and $\Phi$, we write
	\begin{equation}\label{eq:HNN}
		u_h^{\rm NN}
		=
		F\bigl(\Psi(z),\Phi(z)\bigr),
	\end{equation}
	where the representation $F$ is chosen so that
	$\mathcal A u_h^{\rm NN}=0$ holds by construction. The network parameters
	are therefore determined solely from the boundary residual:
	\begin{equation}\label{eq:LOSS_general}
		\mathcal L_{\partial\Omega}
		=
		\frac1N\sum_{i=1}^N
		\left|
		\mathcal B u_h^{\rm NN}(z_i)
		+
		\mathcal B\widetilde u_p(x_i,y_i)
		-
		g(x_i,y_i)
		\right|^2.
	\end{equation}
	Thus, the HPIHNN approximation is given by
	\[
	u^{\rm HPIHNN}
	=
	u_p+u_h^{\rm NN}.
	\]
	
\begin{figure}
		\centering % 图片居中
		\includegraphics[width=0.9\textwidth]{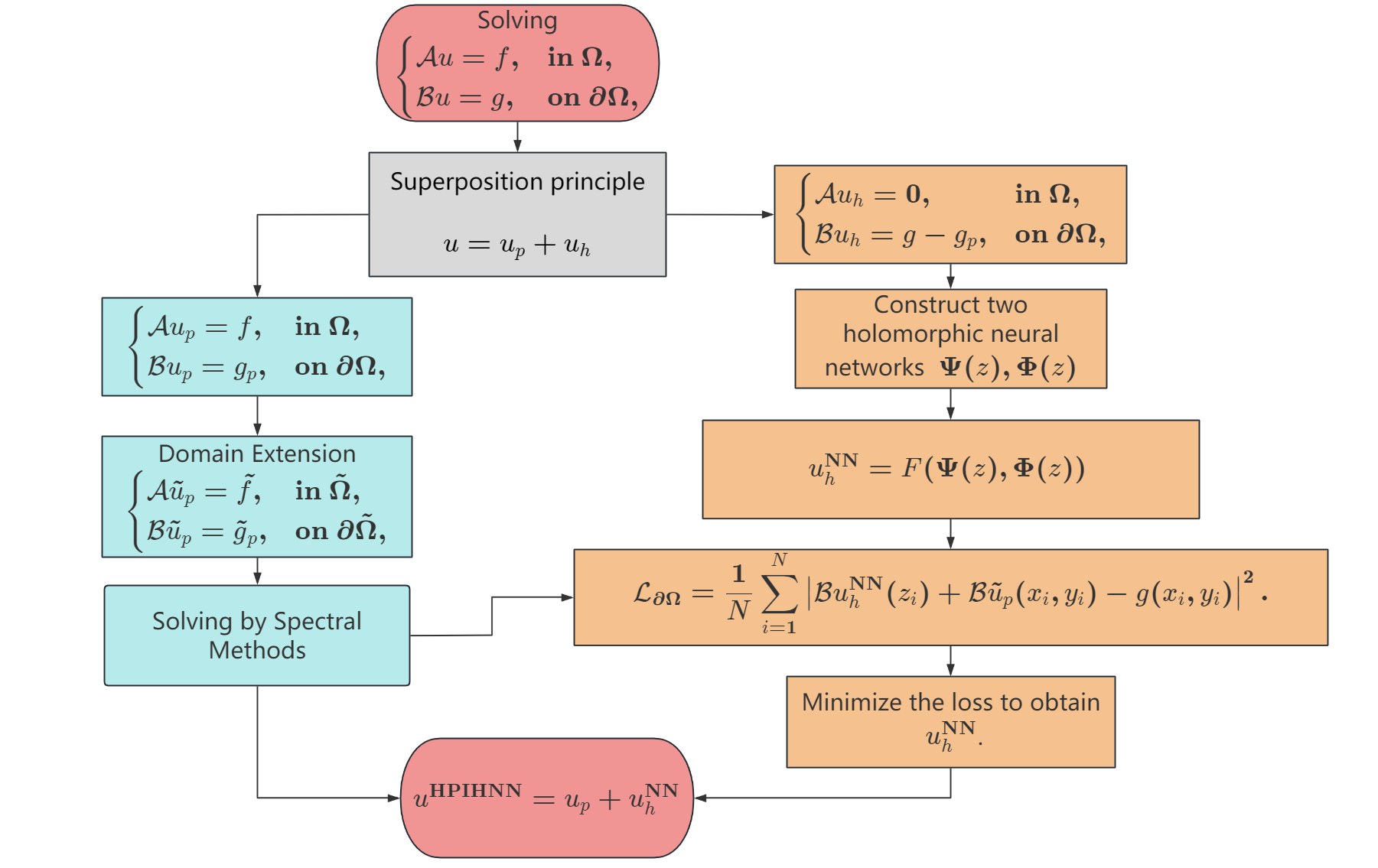}
		\caption{HPIHNN Flowchart}
		\label{fig:Flowchart} % 为图片设置标签，方便引用
	\end{figure}

	% Finally, the solution to the original problem is obtained via the principle of superposition as $u = u_1 + u_{NN}$. As evidenced by this procedure, HPIHNN preserves the inherent advantages of the original PIHNN, its natural proficiency in handling complex geometric boundaries and its high computational efficiency through dimensionality reduction of the sampling space. Simultaneously, it extends the applicability of PIHNNs from strictly homogeneous problems to more general non-homogeneous cases. Furthermore, compared to traditional spectral methods, HPIHNN exhibits superior robustness in handling problems where the solution may lack smoothness despite the source term $f$ being smooth. The computational workflow of HPIHNN is illustrated in the flowchart in Fig. \ref{fig:Flowchart}, and the detailed implementation is summarized in Algorithm \ref{alg:PIHNN-S}.
	This construction retains the boundary-only training structure of PIHNNs and avoids interior collocation for the homogeneous correction, making it well suited to problems on complex geometries. At the same time, the particular--homogeneous decomposition extends the holomorphic neural formulation from homogeneous to nonhomogeneous linear PDEs. Compared with a direct spectral discretization on the original domain, the proposed approach separates the treatment of the source term from that of the geometric boundary, which is advantageous when the solution exhibits reduced regularity despite a smooth source term. The overall HPIHNN workflow is illustrated in Figure~\ref{fig:Flowchart}, and the implementation is summarized in Algorithm~\ref{alg:PIHNN-S}. Section~\ref{sec3.2} gives the holomorphic representations for the homogeneous biharmonic, Stokes, and Helmholtz
	equations and specifies the corresponding operator \(F\) in \eqref{eq:HNN}.
	% \begin{algorithm}
		% 	\caption{HPIHNN for Solving Non-homogeneous Equations}
		% 	\label{alg:PIHNN-S}
		% 	\begin{algorithmic}
			% 		\State \textbf{Step 1:} Decompose the original problem \eqref{eq:PDE1} into two subproblems \eqref{eq:Superposition principle1} and \eqref{eq:Superposition principle2} based on the principle of superposition.
			% 		\State \textbf{Step 2:} Solve the extended non-homogeneous subproblem \eqref{eq:Superposition principle3} using the spectral method to obtain the solution $\tilde{u}_f$.
			% 		\State \textbf{Step 3:} Sample $N$ collocation points on the
			% 		original boundary $\partial\Omega$ and construct the required
			% 		holomorphic neural networks, denoted by $\Phi(z)$ and, when needed,
			% 		$\Psi(z)$, according to the representation of the governing PDE.
			% 		\State \textbf{Step 4:} Employ eqution \eqref{eq:HNN} to obtain the solution $u_{NN}$ that satisfies the homogeneous governing equation, and construct the loss function \eqref{eq:LOSS_general}. The complex derivatives are computed via automatic differentiation.
			% 		\State \textbf{Step 5:} Minimize the loss function \eqref{eq:LOSS_general} to determine the optimized $u_{NN}$.
			% 		\State \textbf{Step 6:} Obtain the final numerical solution $u = \tilde{u}_f\rvert_{\Omega} + u_{NN}$ using the principle of superposition.
			% 	\end{algorithmic}
		% \end{algorithm}	
	
	\begin{algorithm}
		\caption{HPIHNN for nonhomogeneous linear PDEs}
		\label{alg:PIHNN-S}
		\begin{algorithmic}[1]
			\State \textbf{Input:} domain $\Omega$, extended domain $\widetilde{\Omega}$,
			operators $\mathcal A$ and $\mathcal B$, source term $f$, boundary data $g$,
			and number of boundary collocation points $N$.
			
			\State Decompose
			\[
			u=u_p+u_h,
			\]
			with
			\[
			\mathcal A u_p=f,
			\qquad
			\mathcal A u_h=0.
			\]
			
			\State Extend the nonhomogeneous problem to $\widetilde{\Omega}$ and solve
			\eqref{eq:Superposition principle3} by a spectral method to obtain
			$\widetilde u_p$.
			
			\State Restrict the particular solution to the original domain,
			\[
			u_p=\widetilde u_p|_{\Omega},
			\]
			and evaluate the induced boundary data
			\[
			g_p=\left.\mathcal B \widetilde{u}_p\right|_{\partial\Omega}.
			\]
			
			\State Sample $\{(x_i,y_i)\}_{i=1}^N\subset\partial\Omega$, set
			$z_i=x_i+\mathrm{i}y_i$, and construct the holomorphic neural potentials
			required by the corresponding PDE representation.
			
			\State Form the homogeneous neural approximation
			\[
			u_h^{\rm NN}=F\bigl(\Psi(z),\Phi(z)\bigr),
			\]
			which satisfies $\mathcal A u_h^{\rm NN}=0$ by construction.
			
			\State Minimize the boundary loss
			\[
			\mathcal L_{\partial\Omega}
			=
			\frac1N\sum_{i=1}^N
			\left|
			\mathcal B u_h^{\rm NN}(z_i)+\mathcal B \widetilde{u}_p(x_i,y_i)-g(x_i,y_i)
			\right|^2,
			\]
			with the required complex derivatives evaluated by automatic differentiation.
			
			\State \textbf{Output:}
			\[
			u^{\rm HPIHNN}=u_p+u_h^{\rm NN}.
			\]
		\end{algorithmic}
	\end{algorithm}

	% In the following subsection, Section \ref{sec3.2}, we introduce the holomorphic representations for the solutions of the homogeneous Stokes equations and the homogeneous biharmonic equations. Specifically, we provide the explicit forms of the operator $F$ as defined in eqution \eqref{eq:HNN}.
	
	\subsection{Holomorphic representations for selected homogeneous PDEs}
	\label{sec3.2}
	% In addition to the Laplace equations and the homogeneous linear elasticity equations discussed in Section \ref{sec2}, whose holomorphic representations are given by Eqs. \eqref{eq:Lap}, \eqref{eq:KM} and \eqref{eq:Helmholtz_holomorphic_representation} respectively, the solutions to the homogeneous Stokes equations, the homogeneous biharmonic equations and the homogeneous Helmholtz equations can also be expressed in terms of holomorphic functions. These formulations are presented in the following subsections.
	In addition to the Laplace and linear elasticity problems reviewed in Section~\ref{sec2}, several other two-dimensional homogeneous PDEs admit representations in terms of holomorphic functions. In this subsection, we
	summarize the corresponding formulations for the biharmonic, Stokes, and Helmholtz equations.
	
	\subsubsection{Biharmonic equation}
	% We first introduce the holomorphic representation of the fundamental solution for the homogeneous biharmonic equation. Consider the governing equation of the biharmonic equation:
	% \begin{align}
		% 	\nabla^4u=\nabla^2(\nabla^2u)=0.
		% 	\label{eq:Biharmonic0}
		% \end{align}
	% Its general solution can be represented by two holomorphic functions, $\Phi(z)$ and $\Psi(z)$, through the Goursat formula \cite{Goursat}:
	% \begin{equation}
		% 	u(z)=\operatorname{Re}[\bar{z}\Phi(z)+\Psi(z)].
		% 	\label{eq:Biharmonic_basic_solution}
		% \end{equation}
	% In complex analysis, the Laplacian operator can be expressed as :
	% \begin{align}
		% 	\nabla^2=4\frac{\partial}{\partial z\partial\bar{z}},
		% 	\label{eq:complex_value_represent}
		% \end{align}
	% where $\frac{\partial}{\partial z}
	% =\frac{1}{2}(\frac{\partial}{\partial x}-i\frac{\partial}{\partial y})$, $\frac{\partial}{\partial\bar{z}}=\frac{1}{2}(\frac{\partial}{\partial x}+i\frac{\partial}{\partial y})$. 
	
	% \noindent\textbf{Remark:} The function $u(z)=\operatorname{Im}[\bar{z}\Phi(z)+\Psi(z)]$ also satisfies the homogeneous biharmonic equation.\\
	
	Consider
	\begin{equation}\label{eq:Biharmonic0}
		\Delta^2 u=0.
	\end{equation}
	By the Goursat representation \cite{Goursat},
	\begin{equation}\label{eq:Biharmonic_basic_solution}
		u(z)
		=
		\operatorname{Re}\!\left[
		\overline z\,\Phi(z)+\Psi(z)
		\right],
		\qquad
		\Phi,\Psi\in\mathcal H(\Omega_{\mathbb C}),
	\end{equation}
	where $z=x+\mathrm{i}y$. Using
	\[
	\frac{\partial}{\partial z}
	=
	\frac12\left(
	\frac{\partial}{\partial x}
	-\mathrm{i}\frac{\partial}{\partial y}
	\right),
	\qquad
	\frac{\partial}{\partial\overline z}
	=
	\frac12\left(
	\frac{\partial}{\partial x}
	+\mathrm{i}\frac{\partial}{\partial y}
	\right),
	\]
	the Laplacian is
	\begin{equation}\label{eq:complex_value_represent}
		\Delta
		=
		4\frac{\partial^2}{\partial z\,\partial\overline z}.
	\end{equation}
	Hence, \eqref{eq:Biharmonic_basic_solution} satisfies
	\eqref{eq:Biharmonic0} by construction.
	
	\begin{remark}
		The imaginary part
		\[
		u(z)=
		\operatorname{Im}\!\left[
		\overline z\,\Phi(z)+\Psi(z)
		\right]
		\]
		is also biharmonic.
	\end{remark}

	\subsubsection{Stokes equations}	
	Consider the two-dimensional homogeneous Stokes system
	\begin{equation}\label{Stokes_equation}
		\begin{cases}
			-\mu\Delta\bu+\nabla p=0, & \text{in } \Omega,\\
			\nabla\cdot\bu=0, & \text{in } \Omega,
		\end{cases}
	\end{equation}
	where $\bu=(u_x,u_y)$ is the velocity, $p$ is the pressure, and $\mu$ is
	the dynamic viscosity.
	
	Introducing a stream function $\psi$ by
	\begin{equation}\label{eq:psi}
		u_x=\partial_y\psi,
		\qquad
		u_y=-\partial_x\psi,
	\end{equation}
	automatically enforces incompressibility. The Stokes equations further imply
	that $\psi$ is biharmonic, i.e.,
	\[
	\Delta^2\psi=0.
	\]
	Hence, using the Goursat representation,
	\[
	\psi
	=
	\operatorname{Im}
	\!\left[
	\overline z\,\Phi(z)+\Psi(z)
	\right],
	\qquad
	\Phi,\Psi\in\mathcal H(\Omega_{\mathbb C}).
	\]
	
	Defining the complex velocity
	\[
	w:=u_x+\mathrm{i}u_y,
	\]
	the velocity and pressure admit the representation
	\begin{equation}\label{Stokes_equation_result}
		\begin{aligned}
			w
			&=
			-\Phi(z)
			+z\,\overline{\Phi'(z)}
			+\overline{\Psi'(z)},\\
			p
			&=
			4\mu\,\operatorname{Re}\!\bigl(\Phi'(z)\bigr)+C,
		\end{aligned}
	\end{equation}
	where $C$ is an arbitrary pressure constant.
	
	\subsubsection{Helmholtz equation}

	Consider the two-dimensional homogeneous Helmholtz equation
	\begin{equation}\label{eq:homogeneous_Helmholtz}
		(\Delta+\kappa^2)u=0
		\qquad \text{in } \Omega,
	\end{equation}
	where $\kappa$ is the wave number. Unlike harmonic functions, Helmholtz
	solutions are not given directly by the real or imaginary parts of
	holomorphic functions. Instead, they can be generated from holomorphic
	functions through the Vekua--Bergman transformation \cite{Vekua}.
	
	Let $\Phi\in\mathcal H(\Omega_{\mathbb C})$. Taking the reference point
	$z_0=0$, the transformation is
	\begin{equation}\label{eq:Vekua_Bergman_transformation}
		\mathcal V_\kappa[\Phi](z)
		=
		\Phi(z)
		-
		\frac{\kappa |z|}{2}
		\int_0^1
		\Phi(tz)
		\frac{
			J_1\!\left(\kappa |z|\sqrt{1-t}\right)
		}{
			\sqrt{1-t}
		}
		\,\mathrm dt,
	\end{equation}
	where $J_1$ denotes the Bessel function of the first kind of order one.
	The transformed function satisfies
	\begin{equation}\label{eq:Vekua_Helmholtz_equation}
		(\Delta+\kappa^2)\mathcal V_\kappa[\Phi]=0.
	\end{equation}
	Hence, the homogeneous Helmholtz solution used in the present framework is
	represented by
	\begin{equation}\label{eq:Helmholtz_holomorphic_representation}
		u(z)
		=
		\operatorname{Re}\!\left(
		\mathcal V_\kappa[\Phi](z)
		\right),
	\end{equation}
	which satisfies \eqref{eq:homogeneous_Helmholtz} by construction.

	\section{Numerical experiments}
	\label{sec4}
	In this section,  we assess the proposed HPIHNN framework on five classes of nonhomogeneous PDEs: a Poisson problem with a corner singularity, biharmonic and Stokes problems on complex or nonconvex domains, linear elasticity problems involving locking and stress concentration, and a Helmholtz problem on an annular domain. Unless otherwise specified, the activation function is $e^{0.5\mathrm{i}z}$. For comparison, the PINN baseline uses the same hyperparameters as HPIHNN. Since PINNs enforce the governing equations through interior residuals, $2500$ additional interior collocation points are used. For the biharmonic, Stokes, and linear elasticity problems, HPIHNN employs two holomorphic networks, whereas the PINN baseline uses a single network. Its width is therefore doubled so that the number of trainable parameters is comparable to, or larger than, that of HPIHNN.
	
	The numerical settings are summarized in
	Table~\ref{tab:experimental_settings}, where $M$ denotes the spectral
	truncation order in each spatial direction. The learning rate is adjusted
	using PyTorch's \texttt{ReduceLROnPlateau} scheduler. Accuracy is measured by the relative  $L^2$ and $L^\infty$ errors,
	\begin{equation}
		\mathrm{RL}_2
		=
		\frac{\|u^{\mathrm{NN}}-u\|_{L^2(\Omega)}}
		{\|u\|_{L^2(\Omega)}},
		\qquad
		\mathrm{RL}_{\infty}
		=
		\frac{\|u^{\mathrm{NN}}-u\|_{L^{\infty}(\Omega)}}
		{\|u\|_{L^{\infty}(\Omega)}},
	\end{equation}
	where $u^{\mathrm{NN}}$ and $u$ denote the numerical and
	reference solutions, respectively.
	
	\begin{table}[htbp]
		\centering
		\caption{Spectral discretization and neural network configurations for the numerical examples.}
		\label{tab:experimental_settings}
		\renewcommand{\arraystretch}{1.2}
		\resizebox{\textwidth}{!}{%
			\begin{tabular}{@{}lllccccc@{}}
				\toprule
				Problem
				& Spectral domain
				& Spectral basis
				& $M$
				& \multicolumn{2}{c}{Hidden layers $\times$ width}
				& Boundary points
				& Epochs \\
				\cmidrule(lr){5-6}
				& & & & HPIHNN & PINN & & \\
				\midrule
				Poisson
				& $[-1,1]^2$
				& Fourier
				& 20
				& $4\times128$ & $4\times128$
				& 540 & 10,000 \\
				
				Biharmonic
				& $[-1,1]^2$
				& Fourier
				& 20
				& $1\times128$ & $3\times128$\,\textsuperscript{b}
				& 200 & 20,000 \\
				
				Stokes
				& $[0,1]^2$
				& Fourier
				& 64
				& $2\times128$ & $2\times256$
				& 200 & 20,000 \\
				
				Elasticity ($\nu=0.499$)
				& $[0,2\pi]^2$
				& Fourier
				& 64
				& $2\times128$ & $2\times256$
				& 200 & 40,000 \\
				
				Elasticity: square
				& $[-0.5,0.5]^2$
				& Chebyshev\textsuperscript{a}
				& 40
				& $5\times32$ & $5\times128$\,\textsuperscript{c}
				& 400 & 10,000 \\
				
				Elasticity: rail
				& $[-4,12]\times[-2,12]$
				& Chebyshev\textsuperscript{a}
				& 40
				& $5\times32$ & $5\times128$\,\textsuperscript{c}
				& 495 & 20,000 \\
				
				Elasticity: perforated plate
				& $[-2.5,0]\times[0,2.5]$
				& Chebyshev\textsuperscript{a}
				& 40
				& $3\times32$ & $3\times128$\,\textsuperscript{c}
				& 700 & 20,000 \\
				
				Elasticity: perforated plate
				& $[-2.5,2.5]^2$
				& Chebyshev\textsuperscript{a}
				& 40
				& $3\times32$ & $3\times128$\,\textsuperscript{c}
				& 240 & 10,000 \\
				
				Helmholtz
				& $[0,1]^2$
				& Fourier
				& 20
				& $4\times128$ & $4\times128$
				& 351 & 10,000 \\
				\bottomrule
			\end{tabular}%
		}
		
		\medskip
		\begin{minipage}{\textwidth}
			\footnotesize
			%Both the inputs and outputs of the complex-valued neural networks are complex-valued.
			The neural networks take complex-valued inputs and produce complex-valued outputs.\\
			For the Fourier spectral method, periodic boundary conditions are imposed at the boundaries of the extended domain; for the Chebyshev spectral method, homogeneous Dirichlet boundary conditions are imposed at the boundaries.\\
			\textsuperscript{a}%A Chebyshev basis is used because the constant body force cannot be accommodated by the Fourier spectral formulation employed here.
			A Chebyshev basis is adopted, since the constant body-force term is not compatible with the periodic structure underlying the Fourier spectral discretization used here.\\
			\textsuperscript{b}%Three hidden layers are used for the PINN
			%because a single-hidden-layer architecture yielded
			%unsatisfactory accuracy.
			A three-hidden-layer architecture is adopted for the PINN, as the single-hidden-layer configuration did not provide sufficient accuracy.\\
			\textsuperscript{c}%For the more challenging elasticity problems involving stress concentration, the PINN width is increased to four times that of each PIHNN network to achieve comparable accuracy.
			For the more challenging elasticity problems involving stress concentration, the PINN width is increased to four times that of each PIHNN network to ensure a comparable level of accuracy.
		\end{minipage}
	\end{table}

	\subsection{Poisson equation}
	\label{sec4.1}		

	Consider the Poisson problem
	\begin{equation}\label{eq:possion}
		\begin{cases}
			-\Delta u=f, & \text{in } \Omega,\\
			u=g, & \text{on } \partial\Omega,
		\end{cases}
	\end{equation}
	where
	\[
	\Omega=(-1,1)^2\setminus\bigl([0,1]\times[-1,0]\bigr)
	\]
	is an L-shaped domain. We prescribe the 
	%\[
	%f(x,y)=2\pi^2\sin(\pi x)\sin(\pi y),
	%\]
	exact solution
	\begin{equation}\label{eq:possion_exact}
		u(r,\theta)
		=
		r^{2/3}\sin\!\left(\frac{2\theta}{3}\right)
		+2+\sin(\pi x)\sin(\pi y).
	\end{equation}
	The term
	\[
	r^{2/3}\sin\!\left(\frac{2\theta}{3}\right)
	\]
	induces reduced regularity at the re-entrant corner $(0,0)$, where the
	gradient becomes singular. This example therefore provides a test of the
	ability of HPIHNN to approximate solutions with corner singularities.
	
	A four-hidden-layer complex-valued network is employed in this example.
	Since the approximation error is concentrated primarily near the singular
	corner and adjacent boundary segments, additional collocation points are
	placed in these regions. The boundary points are divided into two groups,
	and the training loss is defined by
	\begin{equation}\label{eq:possion_weighted_loss}
		\mathcal L
		=
		w_1\mathcal L_1+w_2\mathcal L_2,
		\qquad
		w_1=1,\quad w_2=5.
	\end{equation}
	The first group consists of uniformly distributed boundary points, whereas
	the second is concentrated near $(0,0)$ and the neighboring corner and
	midpoint regions. A total of $540$ boundary collocation points are used.
	
\begin{figure}
		\centering
		\includegraphics[width=1.0\textwidth]{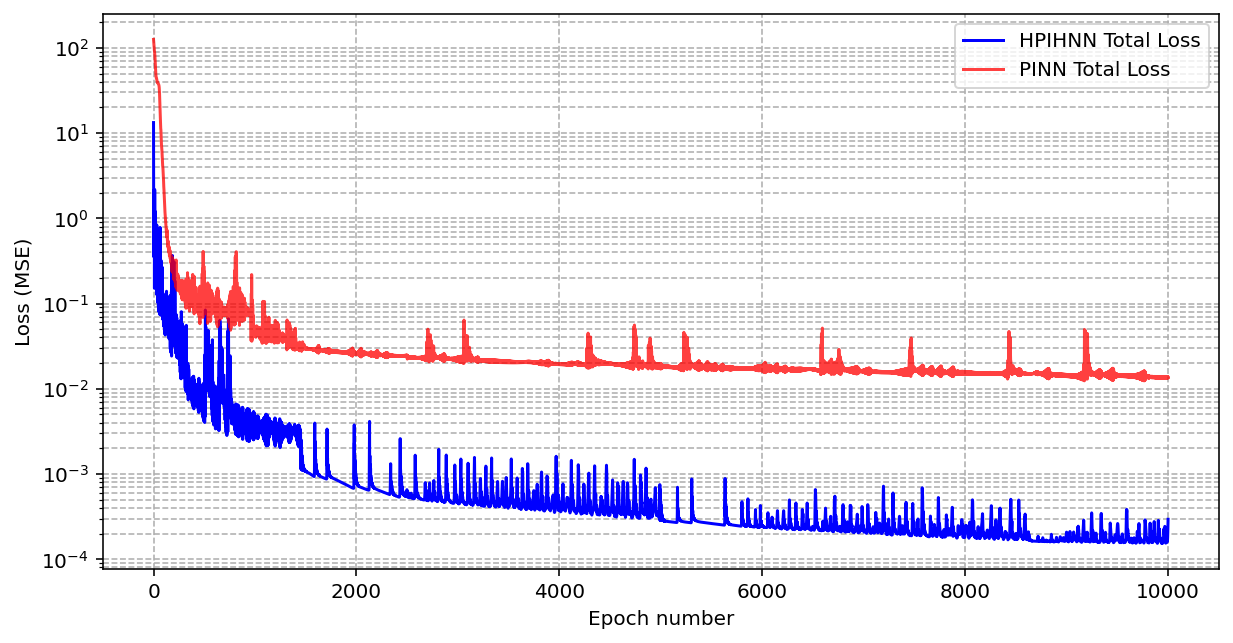}
		\caption{Evolution of the training loss for the HPIHNN.}
		\label{fig:PIHNN_possion_L-shape_loss_spectral}
	\end{figure}

\begin{figure}
		\centering
		\includegraphics[width=0.6\textwidth]{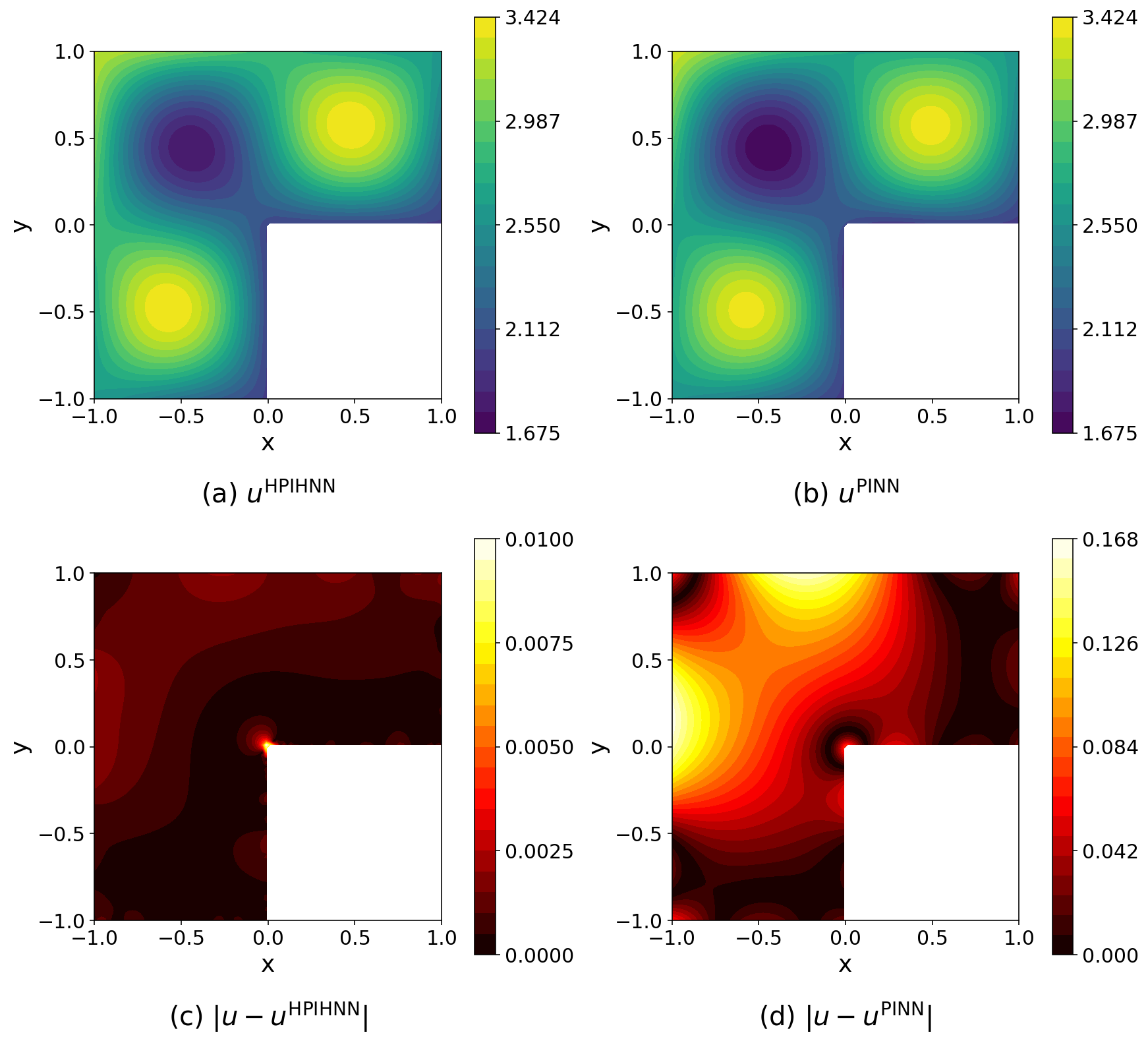}
		\caption{Numerical comparison of HPIHNN and PINN for the Poisson problem \eqref{eq:possion}.}
		\label{fig:PIHNN_possion_L-shape_spectral}
	\end{figure}

	Figure~\ref{fig:PIHNN_possion_L-shape_loss_spectral} shows the training
	history for the homogeneous correction. Figure~\ref{fig:PIHNN_possion_L-shape_spectral}
	compares the HPIHNN and PINN solutions with the reference solution and
	displays the corresponding pointwise errors. The relative errors, numbers
	of trainable parameters, and training times are reported in
	Table~\ref{table:error_results_possion}. For HPIHNN, the dominant errors
	remain localized near the corner singularities, while the interior errors
	are substantially smaller.
	
	\begin{table}[h]
		\centering
		\caption{Accuracy and computational efficiency of HPIHNN and PINN for the Poisson problem \eqref{eq:possion}.}
		\resizebox{10cm}{!}{
			\begin{tabular}{@{}lcccc@{}}
				\toprule
				Method & Computational time& \#parameters & $\mathrm{RL}_2$ & $\mathrm{RL}_{\infty}$\\
				\midrule
				HPIHNN & $57.70\,\mathrm{s}$ & 49,921 
				& $3.01\cdot 10^{-4}$ & $3.54\cdot 10^{-3}$ \\
				PINN & $90.68\,\mathrm{s}$ & 50,049 
				& $1.94\cdot 10^{-2}$ & $4.40\cdot 10^{-2}$ \\
				
				\bottomrule
			\end{tabular}
		}
		\label{table:error_results_possion}
	\end{table}

	\subsection{Biharmonic equation}
	\label{sec4.2}
	We consider the nonhomogeneous biharmonic problem
	\begin{equation}\label{eq:Biharmonic}
		\begin{cases}
			\Delta^2 u=f, & \text{in } \Omega,\\
			u=g_1, & \text{on } \partial\Omega,\\
			\Delta u=g_2, & \text{on } \partial\Omega,
		\end{cases}
	\end{equation}
	posed on a curved five-pointed star domain. Its boundary is parameterized by
	\begin{equation}\label{eq:star}
		\begin{aligned}
			x(\theta)
			&=
			\left[
			R+r-2r\,\operatorname{mod}
			\left(\frac{5\theta}{2\pi},1\right)
			\right]\cos\theta,\\
			y(\theta)
			&=
			\left[
			R+r-2r\,\operatorname{mod}
			\left(\frac{5\theta}{2\pi},1\right)
			\right]\sin\theta,
			\qquad
			\theta\in[0,2\pi],
		\end{aligned}
	\end{equation}
	with $R=0.7$ and $r/R=0.381966$.
	
	The exact solution is prescribed as
	\[
	u(x,y)
	=
	\frac12 x(\sin x\cosh y-\cos x\sinh y)
	+2+\sin(10\pi x)\sin(5\pi y),
	\]
	from which \(f\), \(g_1\), and \(g_2\) are determined.
	%are obtained:
	% \[
	% f
	% =
	% 125^2\pi^4\sin(10\pi x)\sin(5\pi y),
	% \]
	% \[
	% g_1
	% =
	% \frac12 x(\sin x\cosh y-\cos x\sinh y)
	% +2+\sin(10\pi x)\sin(5\pi y),
	% \]
	% and
	% \[
	% g_2
	% =
	% \cos x\cosh y+\sin x\sinh y
	% -125\pi^2\sin(10\pi x)\sin(5\pi y).
	% \]
	
	For the PINN baseline, a network with three hidden layers and $128$ neurons per layer is used. The loss contains separate terms for $u$ and $\Delta u$ on the boundary, with the latter assigned a weight of $5$. Since the source term is much larger in magnitude than the boundary data, the PINN residuals are normalized using an RMS-based relative-residual formulation.
	
\begin{figure}
		\centering
		\includegraphics[width=0.6\textwidth]{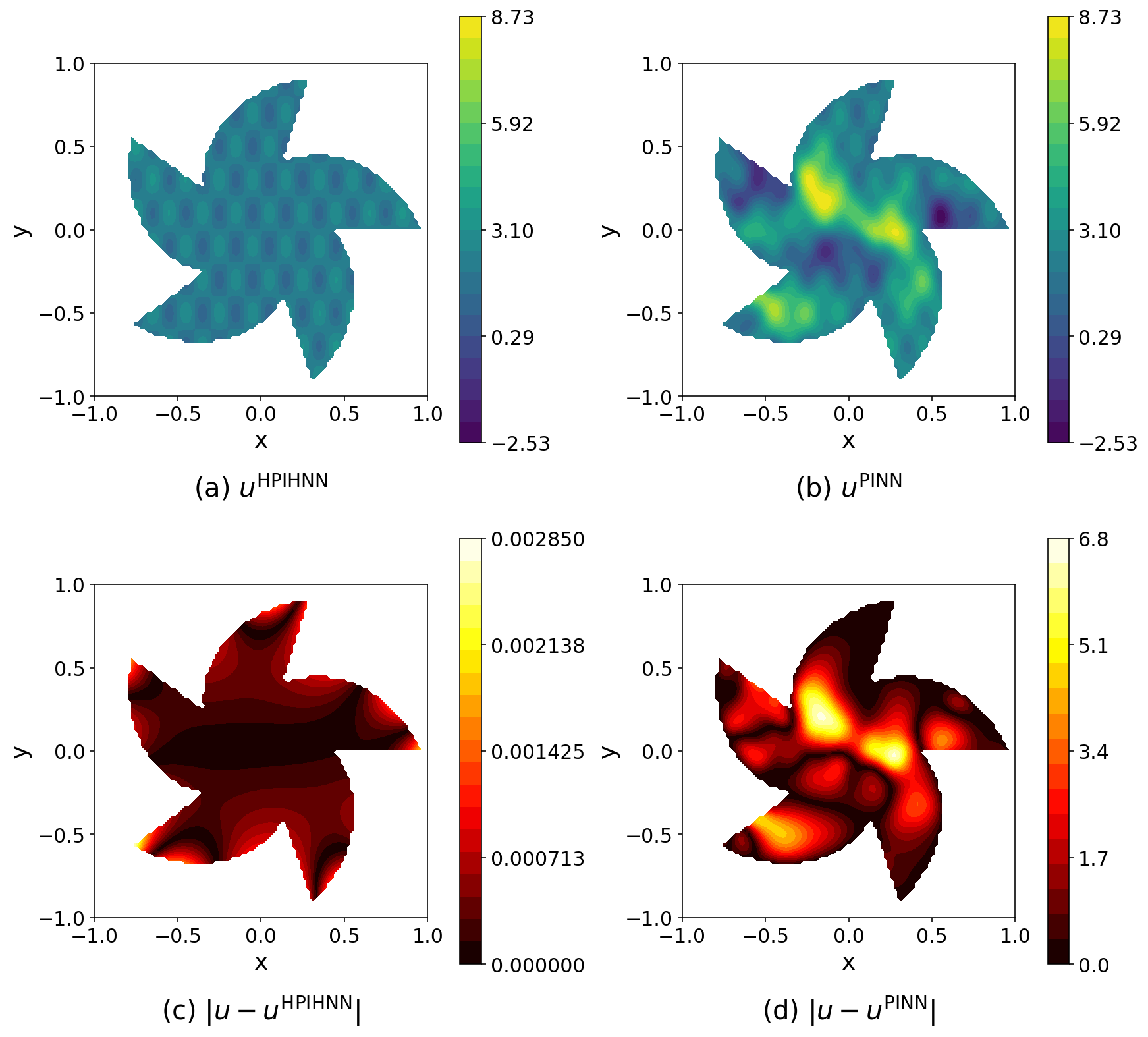}
		\caption{Numerical comparison of HPIHNN and PINN for the biharmonic problem \eqref{eq:Biharmonic}.}
		\label{fig:Biharmonic_spectral_particle}
	\end{figure}

	Figure~\ref{fig:Biharmonic_spectral_particle} compares the numerical and
	exact solutions. The dominant errors are localized near the boundary, while the interior errors are approximately one to two orders of magnitude smaller. The relative errors, numbers of trainable parameters, and training times are
	reported in Table~\ref{table:error_results_biharmonic}. For this
	high-frequency problem on a complex domain, HPIHNN achieves lower errors and
	shorter training times than the PINN baseline.

	\begin{table}[h]
		\centering
		\caption{Accuracy and computational efficiency of HPIHNN and PINN for the biharmonic problem \eqref{eq:Biharmonic}.}
		\resizebox{10cm}{!}{
			\begin{tabular}{@{}lcccc@{}}
				\toprule
				Method & Computational time& \#parameters & $\mathrm{RL}_2$ & $\mathrm{RL}_{\infty}$\\
				\midrule
				HPIHNN & $87.27\,\mathrm{s}$ & 770 
				& $7.93\cdot 10^{-5}$ & $3.41\cdot 10^{-4}$ \\
				PINN & $903.68\,\mathrm{s}$ & 33,537 
				& $2.64\cdot 10^{-1}$ & $4.56\cdot 10^{-1}$ \\
				
				\bottomrule
			\end{tabular}
		}
		\label{table:error_results_biharmonic}
	\end{table}
	
	\subsection{Stokes equation}
	\label{sec4.3}
	% In this section, we consider the Stokes equation with a low viscosity coefficient, as given in Equation~\eqref{eq:stokes}.
	% \begin{align}
		% 	\left\{
		% 	\begin{aligned} 
			% 		-\mu \nabla^2 \mathbf{u} + \nabla p &= \mathbf{f}, &&\text{in }\Omega\\
			% 		\nabla \cdot \mathbf{u} &= 0, &&\text{in } \Omega\\
			% 		\mathbf{u}&=\mathbf{u_0}, &&\text{on } \partial\Omega
			% 	\end{aligned}
		% 	\right.
		% 	\label{eq:stokes}
		% \end{align}
	% where $\Omega$ denotes an L-shaped domain obtained by removing the square subdomain $[0.5,1]\times[0.5,1]$ from the square domain $[0,1]\times[0,1]$. The forcing term is denoted by $\mathbf{f}=(f_x,f_y)$, where
	% $
	% f_x = 2\mu \sin x \cos y - \sin x \cos y,
	% $
	% and
	% $
	% f_y = -2\mu \cos x \sin y - \cos x \sin y.
	% $
	% The viscosity coefficient is chosen as $\mu = 0.001$. The boundary condition is given by $\mathbf{u}_0=(u_0,v_0)$, where
	% $
	% u_0 = 1-y+\sin x \cos y,
	% $
	% and
	% $
	% v_0 = -0.3+x-\cos x \sin y.
	% $
	% Let $\mathbf{u}=(u,v)$ denote the velocity field. Then, the exact solution of the equation is given by
	% \begin{align}
		% 	\begin{cases}
			% 		u=1-y+sinxcosy\\
			% 		v=-0.3+x-cosxsiny\\
			% 		p=cosxcosy
			% 	\end{cases}
		% \end{align}
	% Since the pressure is determined only up to an additive constant,
	% its gauge is fixed by imposing $p(0,0)=1$.
	
	% Comparisons between the numerical and exact solutions are presented in Fig.~\ref{fig:stokes_result_u0.001}. The relative errors, numbers of trainable parameters, and training times are summarized in Table~\ref{table:error_results}.
	In this section, we consider the two-dimensional Stokes problem
	\begin{equation}\label{eq:stokes}
		\begin{cases}
			-\mu\Delta\bu+\nabla p=\mathbf{f}, & \text{in } \Omega,\\
			\nabla\cdot\bu=0, & \text{in } \Omega,\\
			\bu=\bu_0, & \text{on } \partial\Omega,
		\end{cases}
	\end{equation}
	where
	\[
	\Omega=(0,1)^2\setminus([0.5,1]\times[0.5,1])
	\]
	is an L-shaped domain and $\mu=10^{-3}$. The exact solution is prescribed as
	\[
	u=1-y+\sin x\cos y,\qquad
	v=-0.3+x-\cos x\sin y,\qquad
	p=\cos x\cos y,
	\]
	from which the forcing term $\mathbf{f} = (f_x,f_y)^T$ and boundary data
	$\bu_0=(u_0,v_0)^T$ are determined.
	% In particular,
	% \[
	% f_x=2\mu\sin x\cos y-\sin x\cos y,
	% \qquad
	% f_y=-2\mu\cos x\sin y-\cos x\sin y.
	% \]
	Since the pressure is unique only up to an additive constant, we impose
	$p(0,0)=1$ to fix the pressure gauge.
	
\begin{figure}
		\centering
		\includegraphics[width=0.75\textwidth]{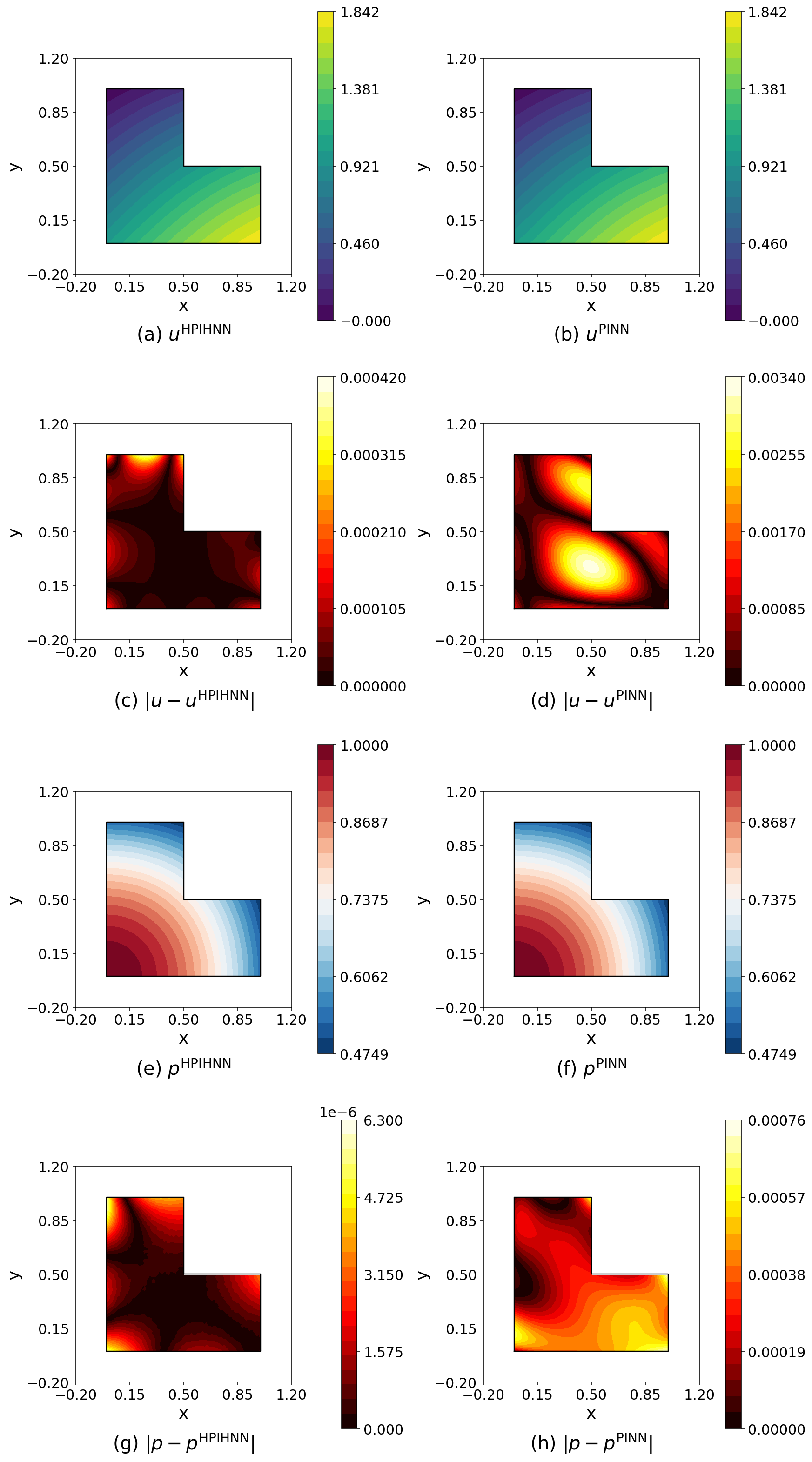}
		\caption{%Exact and numerical solutions of the velocity and pressure fields for the nonhomogeneous Stokes problem on the L-shaped domain.}
		Numerical comparison of HPIHNN and PINN for the Stokes problem \eqref{eq:stokes}.}
	\label{fig:stokes_result_u0.001}
\end{figure}

	Figure~\ref{fig:stokes_result_u0.001} compares the numerical and exact
	solutions, while Table~\ref{table:error_results} reports the relative
	errors, numbers of trainable parameters, and training times.
	
	\begin{table}[h]
		\centering
		\caption{Accuracy and computational efficiency of HPIHNN and PINN for the Stokes problem \eqref{eq:stokes}.}
		\resizebox{\textwidth}{!}{
			\begin{tabular}{@{}lcccccc@{}}
				\toprule
				Method & Computational time& \#parameters & Error norm & $u$ & $v$ & $p$ \\
				\midrule
				\multirow{2}{*}{HPIHNN}
				& \multirow{2}{*}{$61.99\,\mathrm{s}$}
				& \multirow{2}{*}{33794}
				& $\mathrm{RL}_2$
				& $6.28\cdot 10^{-5}$
				& $1.50\cdot 10^{-4}$ 
				& $1.68\cdot 10^{-6}$ 
				\\
				& &
				& 
				$\mathrm{RL}_{\infty}$ 
				& $2.18\cdot 10^{-4}$
				& $3.46\cdot 10^{-4}$ 
				& $6.02\cdot 10^{-6}$ 
				\\
				
				\multirow{2}{*}{PINN}
				& \multirow{2}{*}{$150.57\,\mathrm{s}$}
				& \multirow{2}{*}{67588}
				& $\mathrm{RL}_2$
				& $1.33\cdot 10^{-3}$
				& $3.24\cdot 10^{-3}$ 
				& $4.34\cdot 10^{-4}$ 
				\\
				& &
				& $\mathrm{RL}_{\infty}$ 
				& $1.78\cdot 10^{-3}$
				& $3.43\cdot 10^{-3}$ 
				& $7.48\cdot 10^{-4}$ 
				\\

				\bottomrule
			\end{tabular}
		}
		\label{table:error_results}
	\end{table}
	%The relative velocity error measured in the $L_2$ norm is $5.15\times10^{-5}$, and the relative velocity error measured in the $L_{\infty}$ norm is $2.63\times10^{-4}$. For the pressure field, the absolute error in the $L_2$ norm is $1.43\times10^{-6}$, while the absolute error in the $L_{\infty}$ norm is $4.46\times10^{-6}$. The total computational time is $50.22$ seconds.

\subsection{Linear elasticity equations}
\label{sec4.4}

% In this subsection, we consider several two-dimensional nonhomogeneous linear elasticity problems posed on domains with complex geometries. In the first numerical example, particular attention is given to a nonhomogeneous linear elasticity problem exhibiting the locking phenomenon, where the Poisson ratio approaches $0.5$.

% Locking is a well-known challenge in the numerical simulation of linear elasticity problems. As the Poisson ratio approaches $0.5$, the material becomes nearly incompressible, resulting in a significant disparity between the Lamé parameters, namely $\lambda \gg \mu$, with $\lambda$ tending toward infinity. Under these circumstances, even small numerical errors may be substantially amplified, leading to a noticeable deterioration in the accuracy of the numerical solution. Consequently, the nearly incompressible regime provides a stringent benchmark for assessing the robustness and accuracy of numerical methods.

% Throughout this example, the plane strain assumption is adopted. The governing equations are given by \eqref{eq:elasticity}.

% \begin{equation}
	% 	\left\{
	% 	\begin{aligned} 
		% 		\nabla \cdot \boldsymbol{\sigma} +\mathbf{f}&= \mathbf{0},&&in \quad\Omega \\ 
		% 		\boldsymbol{\sigma} &= 2\mu \boldsymbol{\varepsilon} + \lambda \mathrm{Tr}(\boldsymbol{\varepsilon}) I,&&in \quad\Omega \\ 
		% 		\boldsymbol{\varepsilon} &= \dfrac{\nabla \mathbf{u} + \nabla \mathbf{u}^\top}{2},&&in \quad\Omega \\
		% 		\mathbf{u}&=\mathbf{u_0},&& on \quad \partial\Omega
		% 	\end{aligned}
	% 	\right.
	% 	\label{eq:elasticity}
	% \end{equation}
In this section, we consider several two-dimensional nonhomogeneous linear elasticity
problems on complex domains. The first example focuses on the nearly
incompressible regime, where $\nu\to0.5^-$ and
	$\lambda/\mu\to\infty$, providing a stringent test for numerical methods
susceptible to volumetric locking. Here, $\nu$ denotes Poisson's ratio, and
$\lambda$ denotes the first Lamé parameter. Under the plane-strain assumption, the
governing equations are
\begin{equation}\label{eq:elasticity}
	\begin{cases}
		\nabla\cdot\boldsymbol{\sigma}+\mathbf f=\mathbf 0,
		& \text{in } \Omega,\\
		\boldsymbol{\sigma}
		=
		2\mu\boldsymbol{\varepsilon}
		+\lambda\,\operatorname{tr}(\boldsymbol{\varepsilon})\mathbf I,
		& \text{in } \Omega,\\
		\boldsymbol{\varepsilon}
		=
		\dfrac12(\nabla\mathbf u+\nabla\mathbf u^\top),
		& \text{in } \Omega.
		%\mathbf u=\mathbf u_0.
		%& \text{on } \partial\Omega.
	\end{cases}
\end{equation}

% \subsubsection{A circle with a sector removed}
% \label{sec4.4.1}

% In this example, the computational domain $\Omega$ is a unit disk with the sector corresponding to the angular interval $[0,\pi/4]$ removed. The body force $\mathbf{f}=(f_x,f_y)$ is prescribed as
% \begin{align}
	% 	f_x &= -2(\lambda+2\mu)\sin x\cos y, \\
	% 	f_y &= -2(\lambda+2\mu)\cos x\sin y,
	% \end{align}
% where $\lambda$ and $\mu$ denote the Lamé parameters, which are taken as $\lambda=499$ and $\mu=1$, respectively. These values correspond to a Poisson ratio of $\nu=0.499$.

% The exact displacement and stress fields are given by
% \begin{align}
	% 	\begin{cases}
		% 		u_x = \sin x\cos y + 0.5x,\\[2mm]
		% 		u_y = -0.5\nu y + \cos x\sin y,\\[2mm]
		% 		\sigma_{xx} = 1.525 + 2(\lambda+\mu)\cos x\cos y,\\[2mm]
		% 		\sigma_{yy} = 2(\lambda+\mu)\cos x\cos y,\\[2mm]
		% 		\sigma_{xy} = -2\mu\sin x\sin y,
		% 	\end{cases}
	% 	\label{eq:elasticity_true}
	% \end{align}
% from which the corresponding Dirichlet boundary conditions can be obtained directly.
% Comparisons between the exact and numerical solutions for the displacement magnitude and the stress component $\sigma_{xx}$ are presented in Fig.~\ref{fig:elasticity_result_nu0.499}. 
% The relative $L^2$ errors, numbers of trainable parameters, and training times
% are summarized in Table~\ref{table:nu0.499}.
\subsubsection{A disk with a sector removed}
\label{sec4.4.1}

Let $\Omega$ be the unit disk with the sector
$\theta\in[0,\pi/4]$ removed. We consider the nearly incompressible case
with Lamé parameters
\[
\lambda=499,\qquad \mu=1,
\]
corresponding to $\nu=0.499$. The body force is
\begin{equation}\label{eq:disksector}
	\mathbf f=
	-2(\lambda+2\mu)
	\begin{pmatrix}
		\sin x\cos y\\
		\cos x\sin y
	\end{pmatrix},
\end{equation}
and the exact displacement and stress fields are specified accordingly.
% \begin{equation*}\label{eq:elasticity_true}
	% \begin{aligned}
		% u_x&=\sin x\cos y+\frac12x,\\
		% u_y&=\cos x\sin y-\frac{\nu}{2}y,\\
		% \sigma_{xx}&=1.525+2(\lambda+\mu)\cos x\cos y,\\
		% \sigma_{yy}&=2(\lambda+\mu)\cos x\cos y,\\
		% \sigma_{xy}&=-2\mu\sin x\sin y.
		% \end{aligned}
	% \end{equation*}
Dirichlet data are obtained by restricting the exact displacement to
$\partial\Omega$, i.e., 
\begin{equation}\label{eq:leDich}
	\mathbf u=\mathbf u_0\  \text{on } \partial\Omega.
\end{equation}

\begin{figure}
	\centering
	\includegraphics[width=0.9\textwidth]{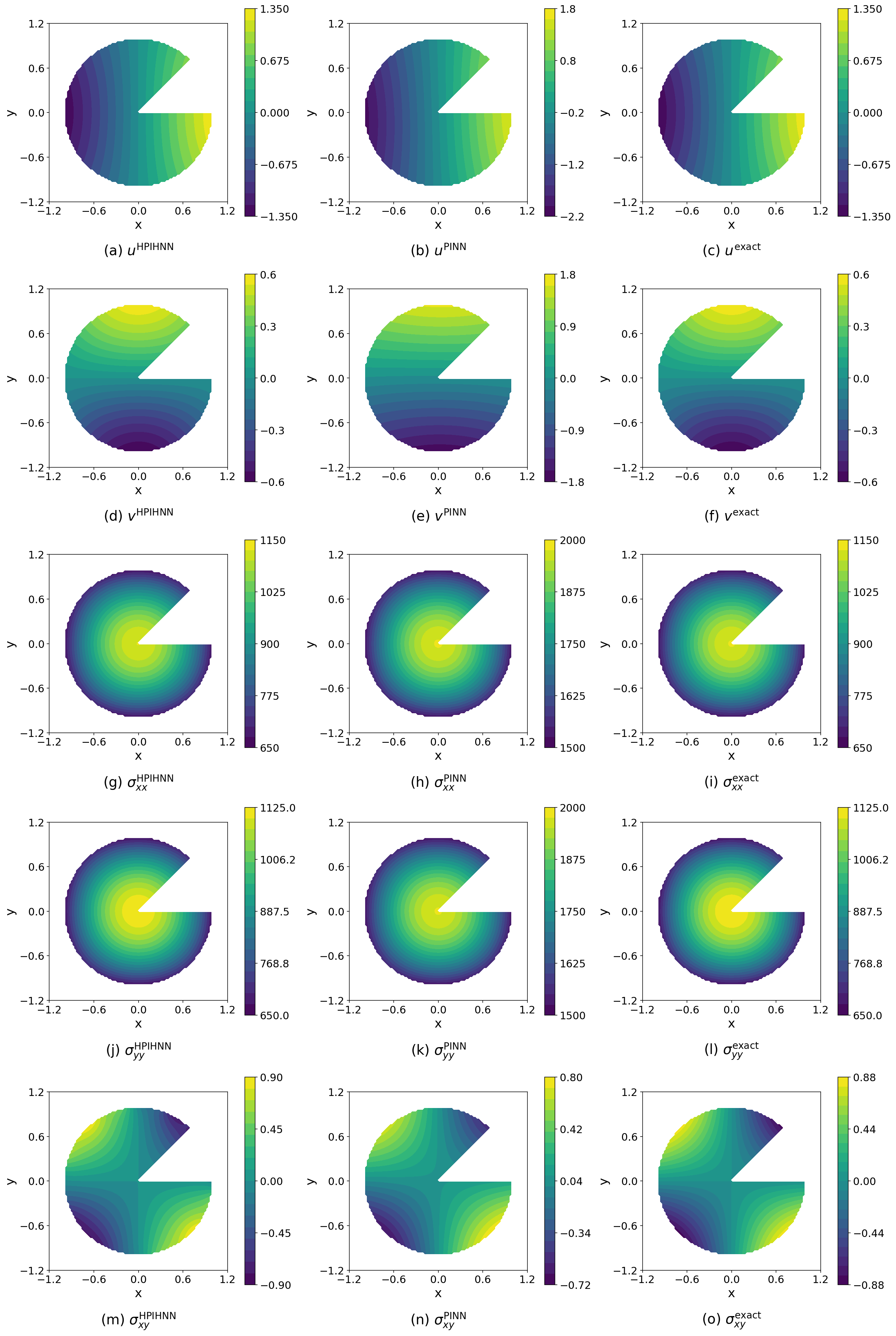}
	\caption{%Comparison between the exact and numerical solutions for the displacement and stress fields for the nearly incompressible nonhomogeneous linear elasticity problem on a unit disk with a sector removed ($\nu=0.499$).
		Numerical comparison of HPIHNN and PINN for the linear elasticity problem \eqref{eq:elasticity}--\eqref{eq:leDich} with $\nu=0.499$.}
	\label{fig:elasticity_result_nu0.499}
\end{figure}

\begin{figure}
	\centering
	\includegraphics[width=0.7\textwidth]{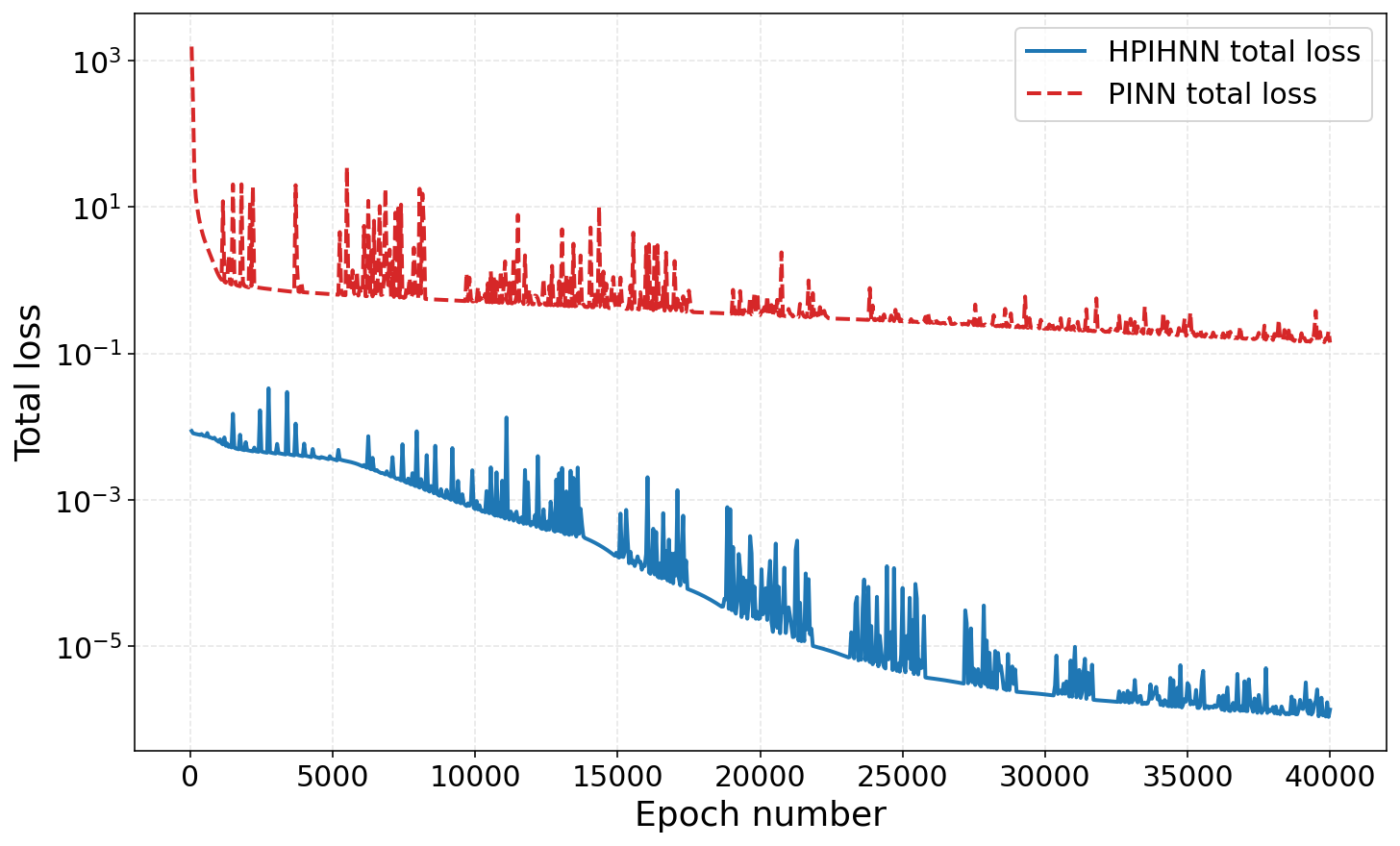}
	\caption{%Comparison between the exact and numerical solutions for the displacement and stress fields for the nearly incompressible nonhomogeneous linear elasticity problem on a unit disk with a sector removed ($\nu=0.499$).
		Numerical comparison of HPIHNN and PINN for the linear elasticity problem \eqref{eq:elasticity}--\eqref{eq:leDich} with $\nu=0.499$.}
	\label{fig:elasticity_result_loss}
\end{figure}

Figure~\ref{fig:elasticity_result_nu0.499} compares the exact and numerical
solutions for the displacement magnitude and $\sigma_{xx}$.
The relative errors, numbers of trainable parameters, and training
times are reported in Table~\ref{table:nu0.499}. Figure~\ref{fig:elasticity_result_loss} compares the training loss histories of HPIHNN and PINN. For this problem involving locking, HPIHNN converges more readily to a lower loss, suggesting improved trainability and approximation accuracy.

\begin{table}[h]
	\centering
	\caption{%Accuracy and performance of HPIHNN and PINN in the second numerical example ($\nu=0.499$) of Section \ref{sec4.4.1}.
		Accuracy and computational efficiency of HPIHNN and PINN for the linear elasticity problem \eqref{eq:elasticity}--\eqref{eq:leDich} with $\nu=0.499$.}
	\resizebox{\textwidth}{!}{
		\begin{tabular}{@{}lcccccccc@{}}
			\toprule
			Method & Computational time & \#parameters & Error norm
			& $\sigma_{xx}$ & $\sigma_{yy}$ & $\sigma_{xy}$ & $u$ & $v$ \\
			\midrule
			\multirow{2}{*}{HPIHNN}
			& \multirow{2}{*}{$111.13\,\mathrm{s}$}
			& \multirow{2}{*}{33794}
			& $\mathrm{RL}_2$
			& $5.42\cdot 10^{-4}$
			& $5.44\cdot 10^{-4}$
			& $2.75\cdot 10^{-2}$
			& $7.52\cdot 10^{-4}$
			& $1.23\cdot 10^{-3}$ \\
			& & &
			$\mathrm{RL}_\infty$
			& $5.20\cdot 10^{-4}$
			& $5.03\cdot 10^{-4}$
			& $2.02\cdot 10^{-1}$
			& $2.27\cdot 10^{-3}$
			& $6.28\cdot 10^{-3}$ \\
			
			\multirow{2}{*}{PINN}
			& \multirow{2}{*}{$266.86\,\mathrm{s}$}
			& \multirow{2}{*}{67074}
			& $\mathrm{RL}_2$
			& $9.01\cdot 10^{-1}$
			& $9.03\cdot 10^{-1}$
			& $5.79\cdot 10^{-1}$
			& $5.70\cdot 10^{-1}$
			& $1.85\cdot 10^{0}$ \\
			& & &
			$\mathrm{RL}_\infty$
			& $7.24\cdot 10^{-1}$
			& $7.26\cdot 10^{-1}$
			& $5.00\cdot 10^{-1}$
			& $5.67\cdot 10^{-1}$
			& $1.92\cdot 10^{0}$ \\
			\bottomrule
		\end{tabular}
	}
	\label{table:nu0.499}
\end{table}

%		 \subsubsection{Realistic, non-trivial geometry}
\subsubsection{Benchmark problems with complex geometries}
\label{engineer}
We next consider three linear elasticity benchmarks with complex geometries
and stress concentrations. The governing system is \eqref{eq:elasticity} with boundary conditions
\begin{equation}\label{eq:elasticity2}
	\begin{cases}
		% \nabla\cdot\boldsymbol{\sigma}+\mathbf f=\mathbf 0,
		% & \text{in } \Omega,\\
		% \boldsymbol{\sigma}
		% =
		% 2\mu\boldsymbol{\varepsilon}
		% +\lambda\,\operatorname{tr}(\boldsymbol{\varepsilon})\mathbf I,
		% & \text{in } \Omega,\\
		% \boldsymbol{\varepsilon}
		% =
		% \dfrac12(\nabla\mathbf u+\nabla\mathbf u^\top),
		% & \text{in } \Omega,\\
		\mathbf u=\mathbf u_0,
		& \text{on } \Gamma_d,\\
		\boldsymbol{\sigma}\mathbf n=\mathbf t,
		& \text{on } \Gamma_f,
	\end{cases}
	\qquad
	\partial\Omega=\Gamma_d\cup\Gamma_f.
\end{equation}
The three benchmark problems share the parameter setting
\[
\mathbf f=(0,-1)^T,\qquad \lambda=\mu=1,
\]
which corresponds to \(\nu=0.25\).

For the first benchmark \cite{PIHNN}, 
%we take
% \begin{equation} 
	% \Omega=[-0.5,0.5]^2,
	% \qquad
	% \mathbf f=(0,-1),
	% \qquad
	% \lambda=\mu=1
	% \quad(\nu=0.25).
	% \end{equation}
a homogeneous Dirichlet condition is imposed on the bottom boundary, a
leftward traction of $1\,\mathrm{MPa}$ on the top boundary, and traction-free
conditions on the lateral boundaries. Since no closed-form solution is
available, a finite element solution computed with constant-strain
triangles on a uniform $256\times256\times2$ mesh is used as reference.

\begin{figure}
	\centering % 图片居中
	\includegraphics[width=0.9\textwidth]{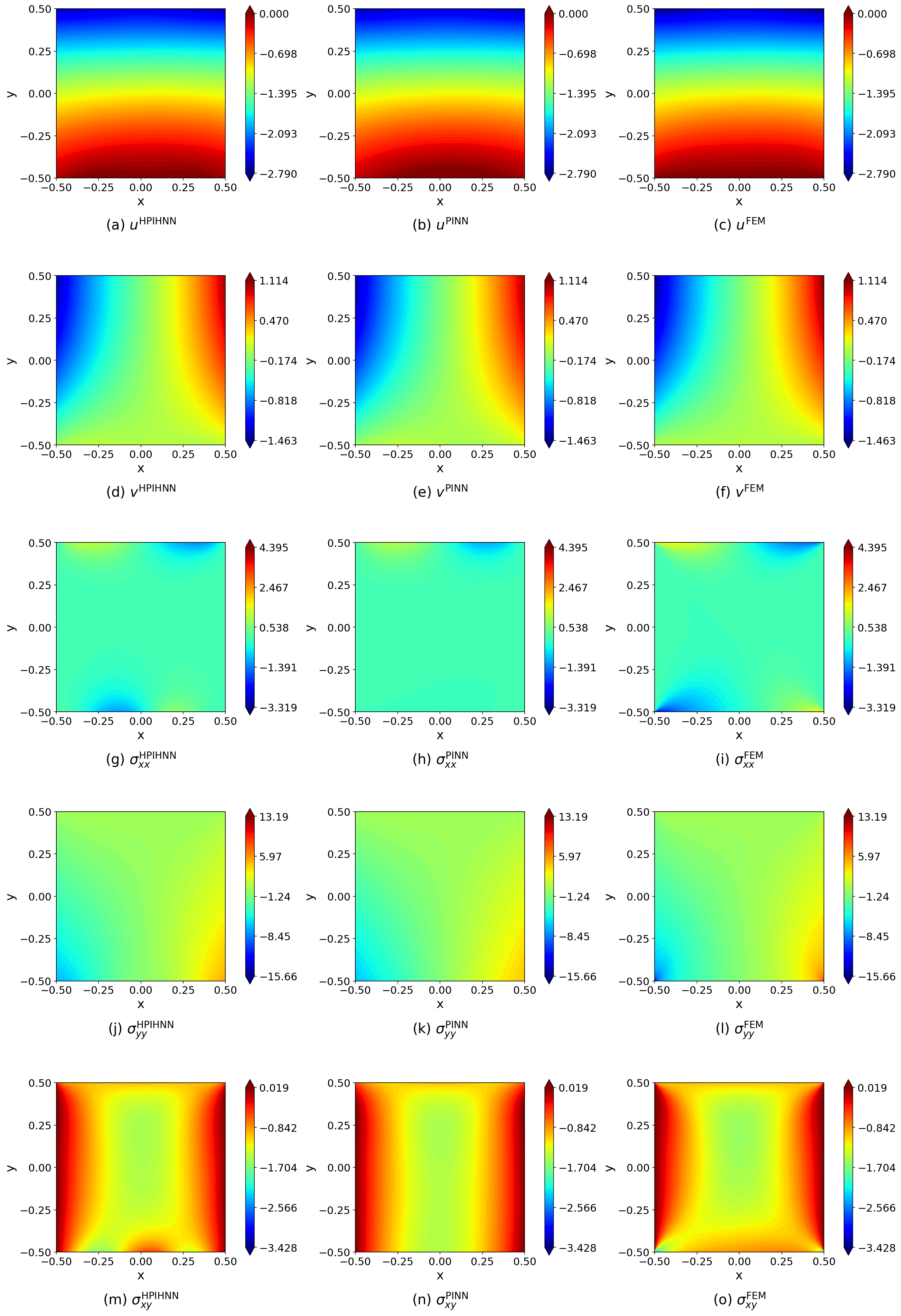}
	\caption{%Comparison of results for an nonhomogeneous Linear Elasticity Problem with Stress Concentration under Leftward Traction
		Numerical comparison of HPIHNN and PINN for the first benchmark of the linear elasticity problems in Section \ref{engineer}.} %\eqref{eq:elasticity}, \eqref{eq:elasticity} and \eqref{eq:leftward}.
	\label{fig:elasticity-complex-geometry1-result} % 为图片设置标签，方便引用
\end{figure}

The solution exhibits stress discontinuities at the upper corners and
corner singularities at the lower corners, with local stress behavior of
approximately $r^{-0.3}$ \cite{example}. For this and the subsequent two
elasticity benchmarks, the particular solution is computed by a Chebyshev
spectral method of order $40$ in each coordinate direction using
Chebyshev--Lobatto points. The PINN baseline uses $10^4$ interior
collocation points, and its hidden-layer width is set to four times that of
each PIHNN network. Figure~\ref{fig:elasticity-complex-geometry1-result} compares the reference
and numerical displacement and stress fields, while
Table~\ref{table:engineer2} reports the relative $L^2$ errors, numbers of
trainable parameters, and training times.

\begin{table}[h]
	\centering
	\caption{%Accuracy and performance of HPIHNN and PINN in the first numerical example of Section \ref{engineer}.
		Accuracy and computational efficiency of HPIHNN and PINN for the first benchmark of the linear elasticity problems %\eqref{eq:elasticity}, \eqref{eq:elasticity2} and \eqref{eq:leftward}.
		in Section \ref{engineer}.}
	\resizebox{\textwidth}{!}{
		\begin{tabular}{@{}lccccccc@{}}
			\toprule
			Method & Computational time & \#parameters & \multicolumn{5}{c}{$\mathrm{RL}_2$} \\
			\cmidrule(lr){4-8}
			& & & $\sigma_{xx}$ & $\sigma_{yy}$ & $\sigma_{xy}$ & $u$ & $v$ \\
			\midrule
			HPIHNN & $255.95\,\mathrm{s}$ & $17284$
			& $5.26\cdot 10^{-1}$
			& $9.31\cdot 10^{-2}$
			& $1.15\cdot 10^{-1}$
			& $1.80\cdot 10^{-2}$
			& $2.72\cdot 10^{-2}$ \\
			
			PINN & $332.29\,\mathrm{s}$ & $66690$
			& $7.55\cdot 10^{-1}$
			& $1.30\cdot 10^{-1}$
			& $1.69\cdot 10^{-1}$
			& $2.38\cdot 10^{-2}$
			& $3.76\cdot 10^{-2}$ \\
			\bottomrule
		\end{tabular}
	}
	\label{table:engineer2}
\end{table}

% The geometry of the second example, shown in
% Fig.~\ref{fig:geometry} \cite{PIHNN}, represents the cross-section of a rail.
% We consider a nonhomogeneous linear elasticity problem defined on the domain
% $\Omega$ associated with this geometry. Based on the example in \cite{PIHNN}, a
% gravitational field is introduced as the source term; that is, the body force is
% set to $\mathbf{f}=(0,-1)$. The Lam\'e constants are chosen as $\lambda=1$ and
% $\mu=1$, corresponding to $\nu=0.25$.

% The boundary conditions are specified as follows: a homogeneous Dirichlet
% condition is imposed on the bottom boundary, a uniform pressure of
% $1\,\mathrm{MPa}$ is applied on the top boundary, and the two lateral
% boundaries are traction-free. Since no exact solution is available, a finite
% element solution is used as the reference solution. The constant-strain
% triangle (CST) method is employed with a mesh size of $0.05$.

\begin{figure}
	\centering % 图片居中
	\includegraphics[width=0.7\textwidth]{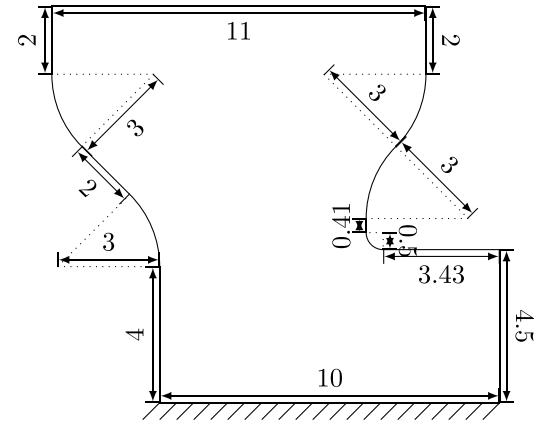}
	\caption{Geometry of the Rail Problem}
	\label{fig:geometry} % 为图片设置标签，方便引用
\end{figure}

The second benchmark, shown in Fig.~\ref{fig:geometry} \cite{PIHNN},
% is posed on a rail-shaped cross-section. We consider the nonhomogeneous
% linear elasticity problem with
% \[
% \mathbf f=(0,-1),\qquad
% \lambda=\mu=1,
% \]
% corresponding to $\nu=0.25$. 
A homogeneous Dirichlet condition is imposed on the bottom boundary, a uniform pressure of $1\,\mathrm{MPa}$ is applied on the top boundary, and the lateral boundaries are traction-free. Since no closed-form solution
is available, a finite element solution computed with constant-strain triangles and mesh size $0.05$ is used as the reference.

Because of the larger spatial extent of the domain, the complex input is
normalized before being passed to the activation function
$e^{0.5\mathrm{i}z}$:
\[
\widehat z=\frac{z-(4+5\mathrm{i})}{8},
\]
which improves the numerical stability of the training process.

\begin{figure}
	\centering % 图片居中
	\includegraphics[width=0.9\textwidth]{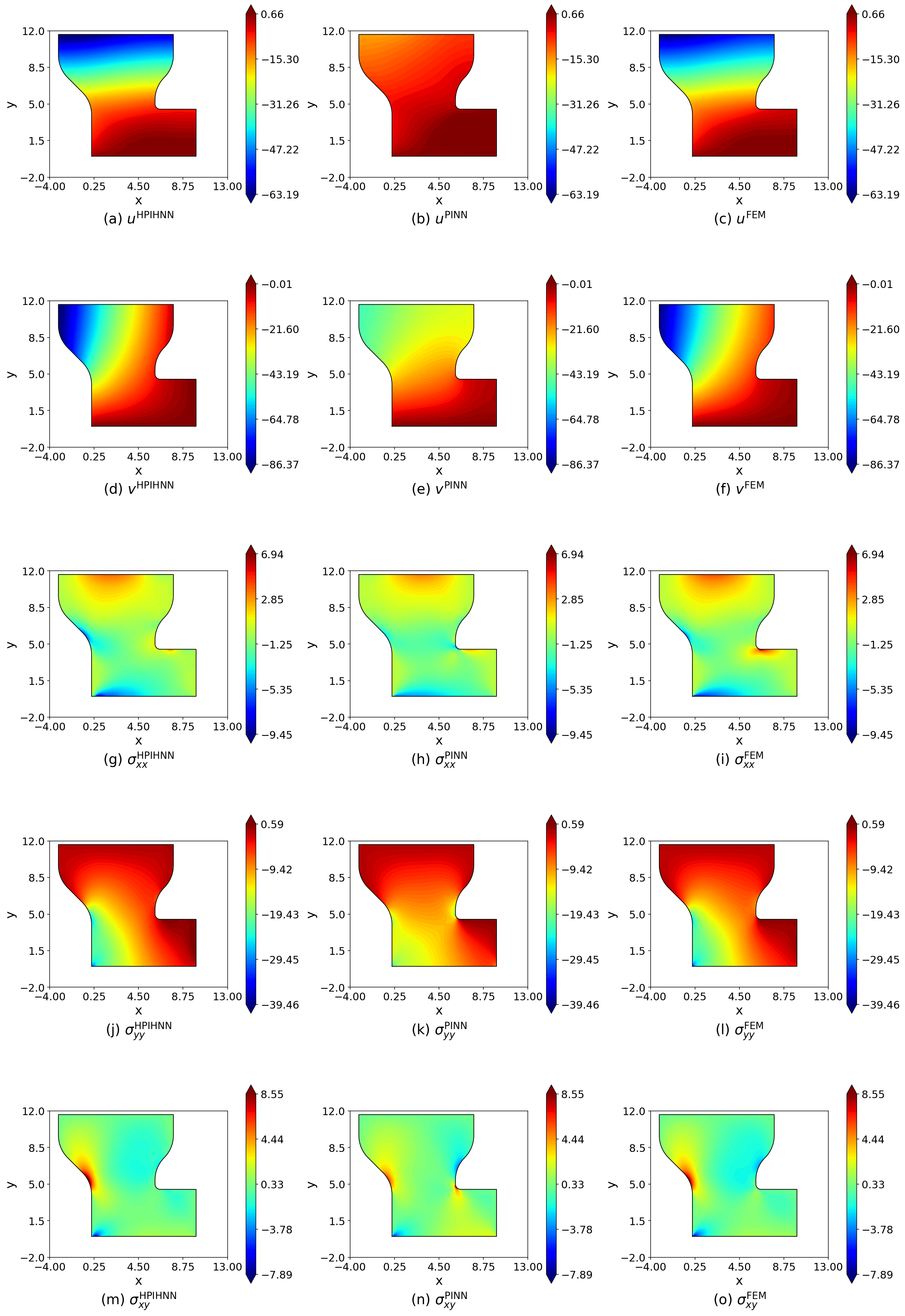}
	\caption{%Comparison of results for the rail stress-concentration problem.
		Numerical comparison of HPIHNN and PINN for the second benchmark (Rail) of the linear elasticity problems in Section \ref{engineer}.}
	\label{fig:elasticity-complex-geometry2-result} % 为图片设置标签，方便引用
\end{figure}

Figure~\ref{fig:elasticity-complex-geometry2-result} compares the reference
and numerical displacement and stress fields, while
Table~\ref{table:engineer1} reports the relative $L^2$ errors, numbers of
trainable parameters, and training times.	

% Moreover, because the computational domain is substantially larger, directly
% using $e^{0.5\mathrm{i}z}$ as the activation function can easily lead to
% instability during training. Therefore, the complex input $z$ is normalized
% with center $4+5\mathrm{i}$ and scaling factor $8$.

% Comparisons between the reference and numerical solutions for the displacement
% components $u$ and $v$, as well as the stress components
% $\sigma_{xx}$, $\sigma_{xy}$, and $\sigma_{yy}$, are shown in
% Fig.~\ref{fig:elasticity-complex-geometry2-result}. The relative $L^2$ errors,
% numbers of trainable parameters, and training times are summarized in
% Table~\ref{table:engineer1}.

\begin{table}[h]
	\centering
	\caption{%Accuracy and performance of HPIHNN and PINN in the second numerical example of Section \ref{engineer}.
		Accuracy and computational efficiency of HPIHNN and PINN for the second benchmark (Rail) of the linear elasticity problems 
		in Section \ref{engineer}.}
	\resizebox{\textwidth}{!}{
		\begin{tabular}{@{}lccccccc@{}}
			\toprule
			Method & Computational time & \#parameters & \multicolumn{5}{c}{$\mathrm{RL}_2$} \\
			\cmidrule(lr){4-8}
			& & & $\sigma_{xx}$ & $\sigma_{yy}$ & $\sigma_{xy}$ & $u$ & $v$ \\
			\midrule
			HPIHNN & $553.32\,\mathrm{s}$ & $17284$
			& $2.61\cdot 10^{-1}$
			& $7.38\cdot 10^{-2}$
			& $2.37\cdot 10^{-1}$
			& $5.15\cdot 10^{-2}$
			& $6.40\cdot 10^{-2}$ \\
			
			PINN & $690.45\,\mathrm{s}$ & $66690$
			& $4.59\cdot 10^{-1}$
			& $3.03\cdot 10^{-1}$
			& $6.38\cdot 10^{-1}$
			& $7.55\cdot 10^{-1}$
			& $3.98\cdot 10^{-1}$ \\
			\bottomrule
		\end{tabular}
	}
	\label{table:engineer1}
\end{table}

\begin{figure}
	\centering % 图片居中
	\includegraphics[width=0.6\textwidth]{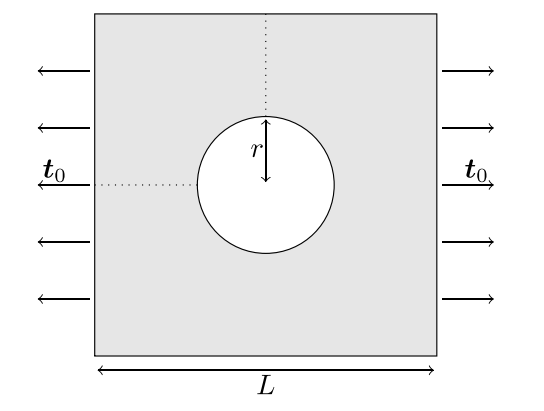}
	\caption{Geometry for the circular-hole stress-concentration problem.}
	\label{fig:elasticity-complex-geometry2} % 为图片设置标签，方便引用
\end{figure}

For the third benchmark \cite{PIHNN}, we consider a square plate of side
length $L=5$ with a central circular hole of radius $r=0.5$ as in Fig.~\ref{fig:elasticity-complex-geometry2}. Uniform tensile
tractions of $1\,\mathrm{MPa}$ are applied on the two lateral boundaries,
while the remaining boundaries are traction-free.
% A body force
% \[
% \mathbf f=(0,-1)
% \]
% is added to obtain a nonhomogeneous problem. The Lam\'e parameters are
% chosen as
% \[
% \lambda=\mu=1,
% \]
% corresponding to $\nu=0.25$. 
By symmetry, only the upper-left quarter of the plate is considered, as
shown in Fig.~\ref{fig:elasticity-complex-geometry2}. On the symmetry
boundaries, we impose
\begin{equation}\label{eq:interface continuity conditions}
	\mathbf u\cdot\mathbf n=0,
	\qquad
	(\boldsymbol{\sigma}\mathbf n)\cdot\mathbf t=0,
\end{equation}
where $\mathbf n$ and $\mathbf t$ denote the unit normal and tangent vectors,
respectively. Since no closed-form solution is available, a finite element
solution computed with constant-strain triangles and a maximum mesh size of
$0.01$ is used as the reference.

To improve training stability on the relatively large domain, the complex
input is normalized as
\[
\widehat z
=
\frac{z-(-1.25+1.25\mathrm{i})}{2.5}.
\]
The PINN baseline uses $10^4$ interior collocation points.

\begin{figure}
	\centering % 图片居中
	\includegraphics[width=0.9\textwidth]{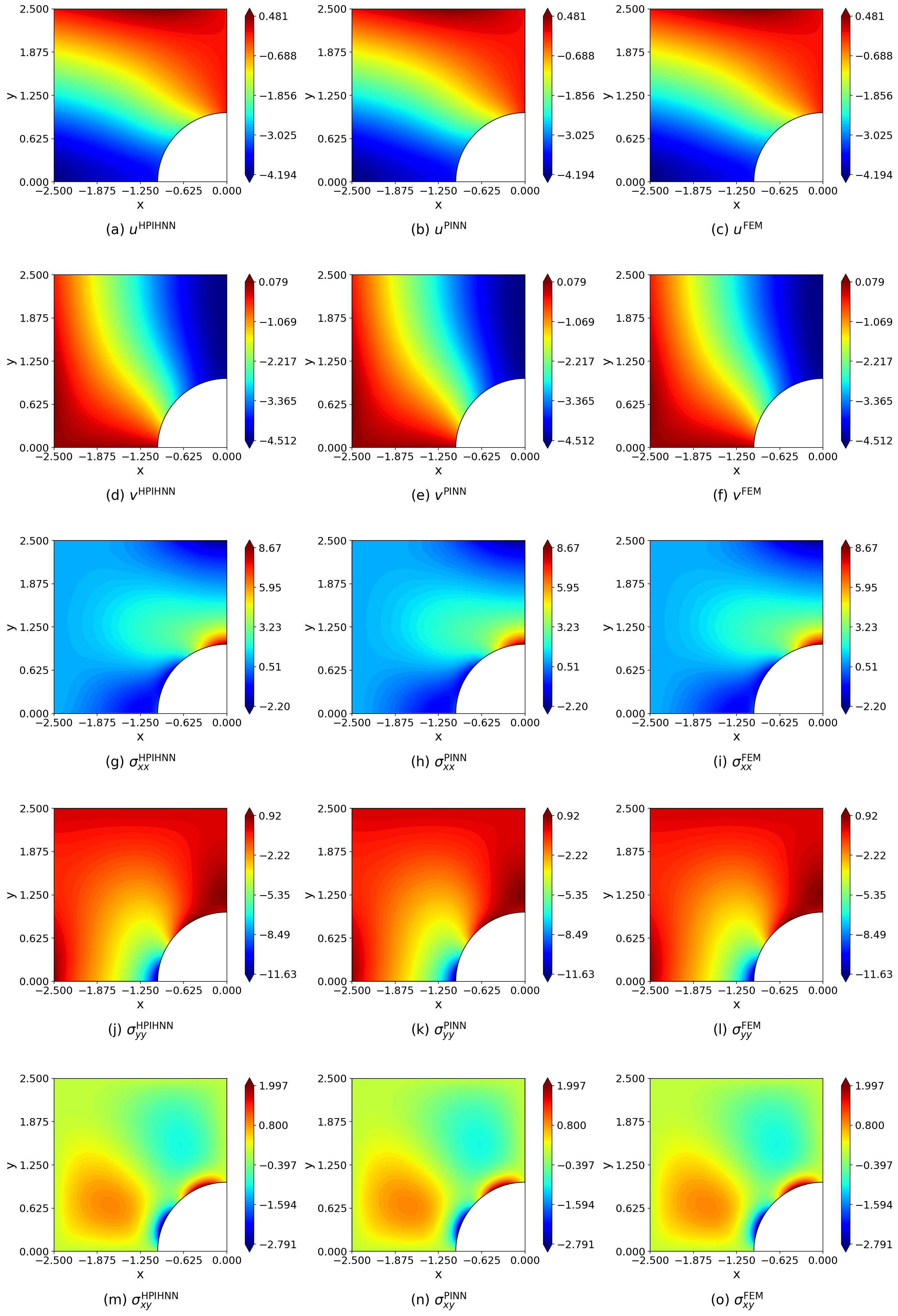}
	\caption{%Comparison of results for a nonhomogeneous linear elasticity problem for the circular-hole stress-concentration problem.
		Numerical comparison of HPIHNN and PINN for the third benchmark  of the linear elasticity problems in Section \ref{engineer}.}
	\label{fig:elasticity-complex-geometry3-result} % 为图片设置标签，方便引用
\end{figure}

Figure~\ref{fig:elasticity-complex-geometry3-result} compares the reference
and numerical displacement and stress fields, while
Table~\ref{table:engineer3} reports the relative $L^2$ errors, numbers of
trainable parameters, and training times.

\begin{table}[h]
	\centering
	\caption{%Accuracy and performance of HPIHNN and PINN in the third numerical example of Section \ref{engineer}.
		Accuracy and computational efficiency of HPIHNN and PINN for the third benchmark of the linear elasticity problems 
		in Section \ref{engineer}.}
	\resizebox{\textwidth}{!}{
		\begin{tabular}{@{}lccccccc@{}}
			\toprule
			Method & Computational time & \#parameters & \multicolumn{5}{c}{$\mathrm{RL}_2$} \\
			\cmidrule(lr){4-8}
			& & & $\sigma_{xx}$ & $\sigma_{yy}$ & $\sigma_{xy}$ & $u$ & $v$ \\
			\midrule
			HPIHNN & $153.38\,\mathrm{s}$ & $8836$
			& $1.20\cdot 10^{-2}$
			& $5.68\cdot 10^{-3}$
			& $2.13\cdot 10^{-2}$
			& $2.42\cdot 10^{-3}$
			& $1.88\cdot 10^{-3}$ \\
			
			PINN & $243.34\,\mathrm{s}$ & $33666$
			& $1.22\cdot 10^{-2}$
			& $6.66\cdot 10^{-3}$
			& $2.18\cdot 10^{-2}$
			& $2.86\cdot 10^{-3}$
			& $1.89\cdot 10^{-3}$ \\
			\bottomrule
		\end{tabular}
	}
	\label{table:engineer3}
\end{table}

\subsubsection{Stress concentration in a heterogeneous domain}

We consider the linear elasticity problem \eqref{eq:elasticity} with geometry shown in Fig.~\ref{fig:elasticity-complex-geometry2}
with body force $\mathbf f=(0,-1)^T.$ Uniform outward tractions of magnitude $1\,\mathrm{MPa}$ are applied on the
left and right boundaries, while the top and hole boundaries are
traction-free. Homogeneous Dirichlet conditions,
\[
u=v=0,
\]
are imposed on the bottom boundary. The domain is partitioned into four equal-area subdomains occupied by distinct isotropic elastic materials. The Lam\'e parameters
$(\lambda,\mu)$, expressed in MPa, are assigned as
\[
(1,1),\qquad (1,0.5),\qquad (0.5,1),\qquad (0.5,0.5), 
\]
in the upper-left, lower-left, upper-right, and lower-right subdomains,
respectively. Perfect bonding is imposed across material interfaces through
continuity of displacement and traction.

Because the material coefficients are piecewise constant, the particular solution is computed separately in each subdomain on a common extended
domain. The PINN baseline uses $10^4$ interior collocation points, while
both methods employ $30$ collocation points on each interface segment. The
same input normalization as in the preceding example is used for HPIHNN.

\begin{figure}
	\centering % 图片居中
	\includegraphics[width=0.9\textwidth]{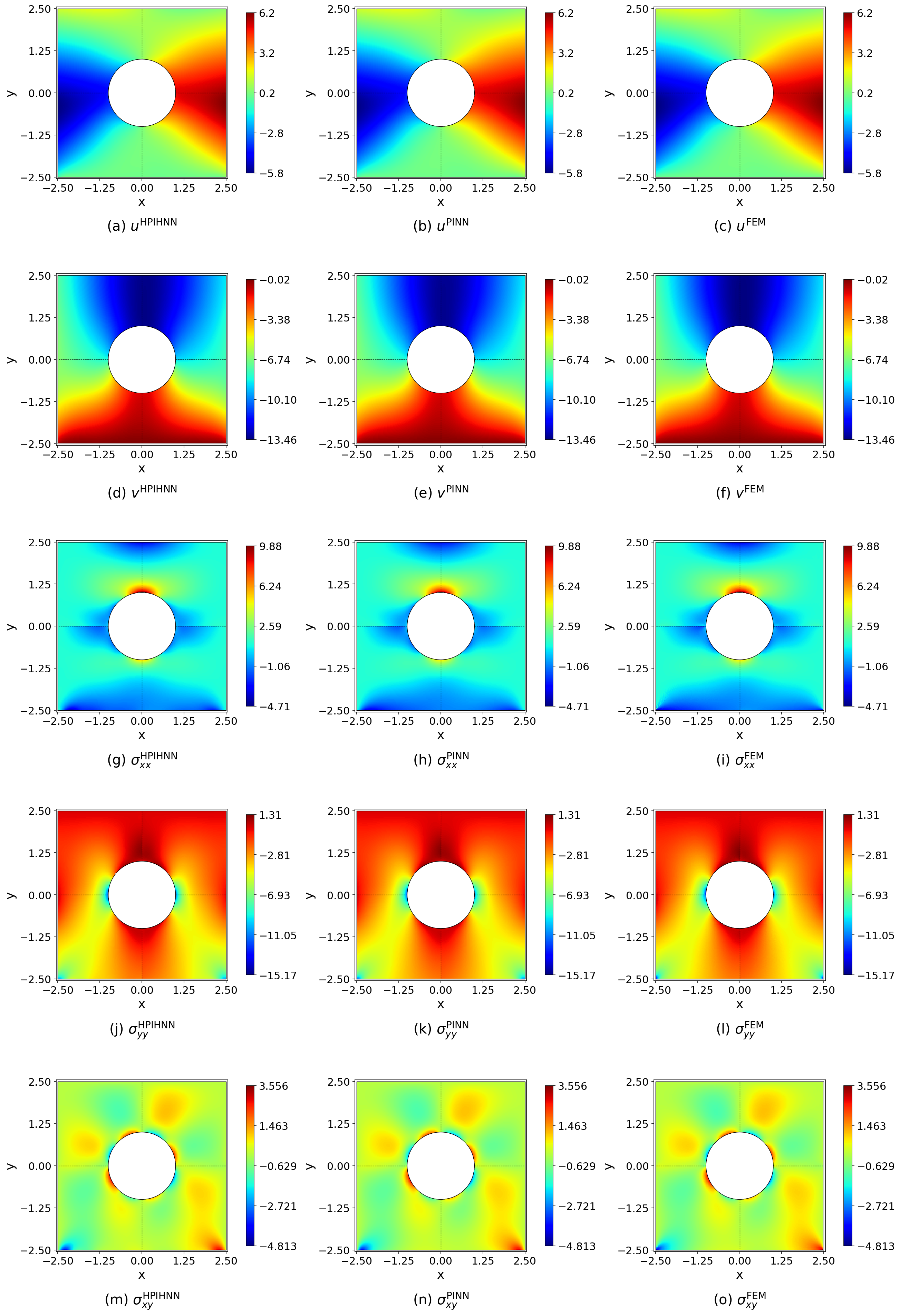}
	\caption{%Comparison of results for a nonhomogeneous linear elasticity problem with material interfaces and stress concentration.
		Numerical comparison of HPIHNN and PINN for the linear elasticity problem 
		in a heterogeneous domain.}
	\label{fig:elasticity-complex-geometry4-result} % 为图片设置标签，方便引用
\end{figure}

Since no closed-form solution is available, a finite element solution
computed with constant-strain triangles and a maximum mesh size of $0.01$
is used as reference. Figure~\ref{fig:elasticity-complex-geometry4-result}
compares the displacement and stress fields. 
%and Fig.~\ref{fig:elasticity-complex-geometry4-loss} shows the corresponding training histories. 
The relative $L^2$ errors, numbers of trainable parameters, and training times are reported in
Table~\ref{table:engineer4}.

% \begin{figure}
	% 	\centering % 图片居中
	% 	\includegraphics[width=1.0\textwidth]{elasticity-realistic4-loss-nonerr.png}
	% 	\caption{Training histories of the individual loss components for HPIHNN and PINN for the linear elasticity problem in a heterogeneous domain.}
	% 	\label{fig:elasticity-complex-geometry4-loss} % 为图片设置标签，方便引用
	% \end{figure}

\begin{table}[h]
	\centering
	\caption{
		%Accuracy and performance of HPIHNN and PINN for the stress concentration problem with material interfaces.
		Accuracy and computational efficiency of HPIHNN and PINN for the linear elasticity problem 
		in a heterogeneous domain.}
	\resizebox{\textwidth}{!}{
		\begin{tabular}{@{}lccccccc@{}}
			\toprule
			Method & Computational time & \#parameters & \multicolumn{5}{c}{$\mathrm{RL}_2$} \\
			\cmidrule(lr){4-8}
			& & & $\sigma_{xx}$ & $\sigma_{yy}$ & $\sigma_{xy}$ & $u$ & $v$ \\
			\midrule
			HPIHNN & $311.54\,\mathrm{s}$ & $35344$
			& $6.78\cdot 10^{-2}$
			& $2.33\cdot 10^{-2}$
			& $1.06\cdot 10^{-1}$
			& $7.65\cdot 10^{-3}$
			& $5.69\cdot 10^{-3}$ \\
			
			PINN & $620.90\,\mathrm{s}$ & $134664$
			& $6.26\cdot 10^{-2}$
			& $2.28\cdot 10^{-2}$
			& $1.01\cdot 10^{-1}$
			& $3.43\cdot 10^{-2}$
			& $1.23\cdot 10^{-2}$ \\
			\bottomrule
		\end{tabular}
	}
	\label{table:engineer4}
\end{table}

% \subsection{Helmholtz equation}
% \label{sec4.5}
% In this subsection, we consider an inhomogeneous Helmholtz problem:
% \begin{equation}
	% 	\left\{
	% 	\begin{aligned}
		% 		\left(\nabla^2+\kappa^2\right)u &= f,
		% 		&& \text{in }\Omega,\\
		% 		u &= u_0,
		% 		&& \text{on }\partial\Omega.
		% 	\end{aligned}
	% 	\right.
	% 	\label{eq:inhomogeneous_Helmholtz_example}
	% \end{equation}
% where $\Omega$ is an annular domain with outer radius $R=1$ and inner radius $r=0.4$. The wave number is set to $\kappa=5\pi$. The exact solution is prescribed as
% \begin{equation}
	% 	u(x,y)=\sin(\pi x)\sin(\pi y)+\cos(\kappa x),
	% \end{equation}
% which yields the source term
% \begin{equation}
	% 	f(x,y)=\left(\kappa^2-2\pi^2\right)\sin(\pi x)\sin(\pi y).
	% \end{equation}
% The boundary condition $u_0$ is obtained by restricting the exact solution to $\partial\Omega$.

% Comparisons between the numerical and exact solutions are presented in Fig.~\ref{fig:helmholtz_result}. The relative errors, numbers of trainable parameters, and training times are summarized in Table~\ref{table:error_results_helmholtz}.

\subsection{Helmholtz equation}
\label{sec4.5}
We consider the nonhomogeneous Helmholtz problem
\begin{equation}\label{eq:inhomogeneous_Helmholtz_example}
	\begin{cases}
		(\Delta+\kappa^2)u=f, & \text{in } \Omega,\\
		u=u_0, & \text{on } \partial\Omega,
	\end{cases}
\end{equation}
where $\Omega$ is the annulus with inner radius $0.4$ and outer radius $1$,
and $\kappa=5\pi$. We prescribe the exact solution
\begin{equation}
	u(x,y)=\sin(\pi x)\sin(\pi y)+\cos(\kappa x),
\end{equation}
from which the boundary data $u_0$ and source term $f$ are determined.

\begin{figure}
	\centering
	\includegraphics[width=0.6\textwidth]{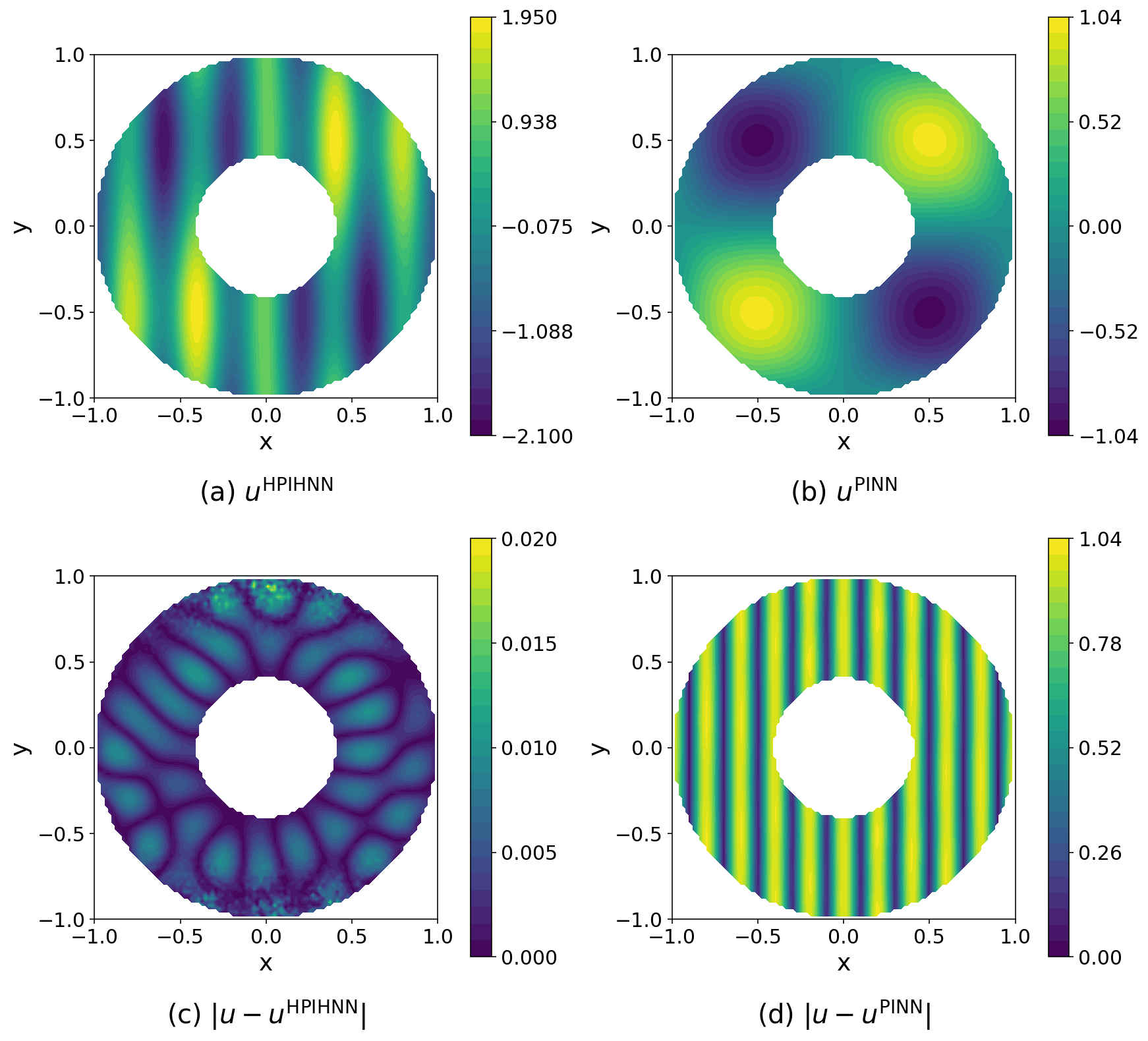}
	\caption{%Exact and numerical solutions of the nonhomogeneous Helmholtz equation on an annular domain.
		Numerical comparison of HPIHNN and PINN for the Helmholtz problem \eqref{eq:inhomogeneous_Helmholtz_example}.}
	\label{fig:helmholtz_result}
\end{figure}

Figure~\ref{fig:helmholtz_result} compares the numerical and exact
solutions, while Table~\ref{table:error_results_helmholtz} reports the
relative errors, numbers of trainable parameters, and training times.	
\begin{table}[h]
	\centering
	\caption{%Accuracy and computational efficiency of HPIHNN and PINN in the Helmholtz numerical example.
		Accuracy and computational efficiency of HPIHNN and PINN for  the Helmholtz problem \eqref{eq:inhomogeneous_Helmholtz_example}.}
	\resizebox{9cm}{!}{
		\begin{tabular}{@{}lccc@{}}
			\toprule
			Method & Computational time& \#parameters & $\mathrm{RL}_2$ \\
			\midrule
			HPIHNN
			& $42.06\,\mathrm{s}$
			& 49921
			&$4.90\cdot 10^{-3}$\\
			
			PINN
			&$60.25\,\mathrm{s}$
			&50049
			& $7.70\cdot 10^{-1}$
			\\

			\bottomrule
		\end{tabular}
	}
	\label{table:error_results_helmholtz}
\end{table}
%The relative velocity error measured in the $L_2$ norm is $5.15\times10^{-5}$, and the relative velocity error measured in the $L_{\infty}$ norm is $2.63\times10^{-4}$. For the pressure field, the absolute error in the $L_2$ norm is $1.43\times10^{-6}$, while the absolute error in the $L_{\infty}$ norm is $4.46\times10^{-6}$. The total computational time is $50.22$ seconds.

%=============================================================================		
\section{Conclusion}
\label{sec5}

In present work, we propose HPIHNN, a hybrid framework for nonhomogeneous linear PDEs that combines spectral particular solutions with holomorphic neural approximations of homogeneous corrections. By solving the source-driven component on a regular extended domain and restricting neural-network training to boundary residuals, the method retains the geometric flexibility of PIHNNs while extending them to nonhomogeneous problems.

Numerical experiments on Poisson, biharmonic, Stokes, linear elasticity, and Helmholtz equations demonstrate the effectiveness of the approach for complex geometries, singular or high-frequency solutions, nearly incompressible regimes, and heterogeneous materials. In the considered tests, HPIHNN generally yields lower errors and shorter training times than the PINN baseline.
The present framework is limited to two-dimensional problems admitting suitable holomorphic representations. Future work will focus on extensions to three-dimensional, variable-coefficient, and nonlinear PDEs, as well as PDE-constrained optimization problems.

\section*{Acknowledgment}
This research is partially sponsored by the National Natural Science Foundation of China (No.12371434, No. 12531015, No. 12671515, No. U25A20200), the National Key R \& D Program of China (No.2022YFE03040002), Sichuan University Research Project on Teaching Reform in Graduate Education (No. SCYZTD2025003). 
\section*{Data Availability Statement}
The data that support the findings of this study are available from the corresponding author upon reasonable request.
\section*{Conflict of interest}
The authors declare that they have no conflict of interest.

%\appendix
%\section{Example Appendix Section}
%\label{app1}


%%       \cite{<label>} ==> [1]

%%
%Example citation, See \cite{lamport94}.




\begin{comment}

\end{comment}

\bibliographystyle{elsarticle-num}
\bibliography{references}
\end{document}